# LLM-Assisted Multi-Fidelity Bayesian Optimization: Fusing Expert Knowledge with Numerical Data

Dongjun Han[a], Mingyu Lee[a,*], Ikjin Lee[b,c,*]

[a] Department of Mechanical Engineering, Korea Advanced Institute of Science and Technology (KAIST), Daejeon 34141, Republic of Korea

[b] Department of AX & Mechanical Engineering, Korea Advanced Institute of Science and Technology (KAIST), Daejeon 34141, Republic of Korea

[c] KAIST InnoCORE PRISM-AI Center, Korea Advanced Institute of Science and Technology (KAIST), Daejeon 34141, Republic of Korea

[*] Corresponding authors.

E-mail addresses: ikjin.lee@kaist.ac.kr (I. Lee), mglee91@kaist.ac.kr (M. Lee).

## Abstract

Multi-fidelity Bayesian optimization (MFBO) reduces engineering design cost by combining inexpensive low-fidelity (LF) models with a few accurate high-fidelity (HF) evaluations and sequentially selecting designs to evaluate. However, when HF data are scarce, the LF–HF discrepancy is difficult to learn from numerical data alone, especially when physics omitted in the LF model makes it design-dependent. Yet even without sufficient data, domain experts often know which physics the LF model omits and under which conditions its effect grows. Existing multi-fidelity (MF) models cannot use such qualitative knowledge directly, but large language models (LLMs) can interpret it and compare designs accordingly. The proposed LLM-Assisted MFBO adds a knowledge basis to the conventional MF model: from an expert description of the omitted physics, an LLM judges which of two designs requires the larger signed residual correction, and a pairwise Gaussian process converts these judgments into the basis. HF data calibrate its contribution and sign, so the knowledge is not trusted blindly. On a lithium-ion battery design problem, the proposed method reduced the regret area under the curve of conventional MFBO by 61%. Knowledge quality, rather than the choice of LLM, determined whether the search was accelerated, as shown by a controlled study that varied the two independently. A knowledge-level ablation further quantified the contribution of each knowledge statement. These results show that LLM-interpreted expert knowledge can complement numerical data as an information source in MFBO.

## 1. Introduction

Engineering design optimization increasingly relies on detailed physics-based simulations in structural, fluid, thermal, and electrochemical applications. Such simulations predict design performance reliably, but a single run can take minutes to hours, which makes conventional optimization methods that require hundreds to thousands of evaluations impractical [1,2]. Surrogate models, which approximate the response from a small number of simulations, and Bayesian optimization (BO), which uses the predictive mean and uncertainty of a surrogate to select the next evaluation sequentially, have therefore been widely adopted [3–5]. A representative strategy for reducing the simulation cost further is multi-fidelity (MF) modeling, which combines an accurate but expensive high-fidelity (HF) model with an inexpensive but less accurate low-fidelity (LF) model that simplifies the underlying physics [6–9]. By capturing the global trend of the response from abundant LF data and correcting the difference between fidelities with limited HF data, MF surrogates and their combination with BO (MFBO) achieve higher efficiency than single-fidelity methods under the same HF budget [10–13].

Although MF methods such as co-Kriging, hierarchical Kriging, and nonlinear information fusion differ in form, they share one feature: the LF–HF relation is inferred from numerical observations alone [14–16]. When the LF model omits a physical phenomenon, the discrepancy varies locally and nonlinearly with the design variables and operating conditions, and learning such a *design-dependent discrepancy* requires sufficient HF observations where the discrepancy grows [9,17]. More fundamentally, studies on model calibration have shown that a scale factor and a discrepancy function cannot be distinguished from a few observations, and that without prior information on the discrepancy the estimates may converge to physically meaningless values [18,19]. In the early stage of MFBO, where HF data are scarcest, performance therefore depends on where prior information on the discrepancy can be obtained. Such information does exist. Even without knowing the exact HF response, a domain expert knows which physical phenomena the LF model omits and under which conditions their influence grows. The difficulty is that this knowledge exists not as a numerical prior or a governing equation but as a natural-language description with intertwined conditions. Methods that place a prior on the location of the optimum in the acquisition function [20] and physics-informed learning that imposes governing equations on the model [21,22] require knowledge in numerical or equation form, and thus cannot bring such descriptions into the discrepancy structure. What is needed is an interpretation step that reads the expert description and translates it into quantitative information usable by an MF model. This step has traditionally been carried out by people through expert elicitation, but translating qualitative knowledge with intertwined conditions into numbers is time-consuming, and the outcome is sensitive to the elicitation procedure [23].

Recent large language models (LLMs) can read domain descriptions and reason conditionally about a given design, which makes them a promising means of automating this interpretation step. LLMs are already used as scientific agents that plan experiments and operate tools [24,25] and as candidate generators in search algorithms [26], and studies that combine LLMs with BO are growing. LLAMBO [27] provides the observation history as natural-language context so that the LLM directly predicts the objective value and its uncertainty at candidate points, and LABO [28] treats numerical predictions from the LLM as an LF information source and combines them with actual observations. For engineering responses whose scales and units differ from problem to problem, however, it is difficult to know in advance how far such numerical outputs of an LLM can be trusted. Indeed, absolute numerical judgments by LLMs have been reported to vary easily with the query format and context, whereas asking an LLM to compare two items yields more consistent judgments [29,30]. Other approaches use qualitative information instead of numbers. LGBO [31] shifts the mean of the Gaussian process (GP) surrogate toward regions that the LLM judges promising, and LILO [32] converts natural-language feedback from a decision maker into pairwise preferences to learn a latent utility function, showing that pairwise comparisons provide more reliable information than scalar utilities. Nevertheless, what these methods ask the LLM still concerns the objective itself, such as where the optimum lies or which outcome is better. In engineering design, what an expert can state with confidence is usually not the location of the optimal design or the ranking of performance, but which physics the LF model omits and where its influence grows. When a physics-based LF model is available, it is therefore more appropriate to ask the LLM a relative question grounded in expert knowledge, "Where does the LF model err more?", rather than to entrust it with judgments on the objective, and to let HF data determine how far this information should be trusted.

This study therefore changes the question posed to the LLM and proposes LLM-Assisted MFBO, which integrates expert knowledge with numerical data in an MF surrogate. Once an expert describes the physical differences between the LF and HF models in natural language, the LLM, instead of predicting numbers, judges only which design in an actively selected pair requires the larger signed residual correction. A pairwise GP converts a small number of such comparisons into a continuous knowledge basis that enters the MF model together with the LF basis, and HF data calibrate its sign and contribution. The LF model is thus retained, and expert knowledge is not trusted blindly but reflected only as far as HF data support it. The contributions of this study are as follows.

- **Expert knowledge is formalized as an information source of the MF surrogate.** Knowledge structured by the LLM is added to the conventional MF model as a knowledge basis that explains the LF–HF discrepancy, and an information-source ablation shows that this basis provides information complementary to LF data and prevents misattribution in the LF coefficient.

- **Knowledge, not the LLM, is shown to govern performance.** In a controlled factorial study of four knowledge qualities and four LLMs, about 93% of the between-condition variance in performance stemmed from the knowledge and about 2% from the LLM.
- **The contribution of each knowledge statement is provided as design information.** A knowledge-level ablation quantifies how much each knowledge statement contributes to optimization performance, indicating which knowledge matters most for the design problem and which should be refined first.
- **Without knowledge, the method reduces exactly to conventional MFBO.** When there is no basis for a judgment, the LLM abstains and the proposed method becomes identical to conventional MFBO; when the direction of the knowledge is globally reversed, HF calibration corrects it through the sign of the coefficient.

The remainder of this paper is organized as follows. Section 2 reviews previous studies on MF modeling, MFBO, and LLM-based BO. Section 3 describes the motivation, key concept, detailed strategies, and overall procedure of the proposed method. Section 4 validates the method on a lithium-ion battery design problem with electrochemical simulations, and Section 5 presents conclusions and future research directions.

## 2. Literature review

The need for this study becomes clear when two lines of research, MF modeling and LLM-based BO, are examined together. Section 2.1 reviews MF surrogate modeling and MFBO. Section 2.2 compares LLM-based BO methods according to the role played by the LLM and positions this study among them.

### 2.1 MF modeling and BO

The goal of MF modeling is to perform prediction and optimization while minimizing the number of accurate but computationally expensive HF evaluations. The probabilistic formulation that represents the relation between fidelity levels with GPs was introduced by Kennedy and O'Hagan [6], and it has become the foundation of co-Kriging, hierarchical Kriging, and various discrepancy-based MF surrogates [14–16,33]. The use of surrogate-based MF methods in engineering design was reported by Forrester et al. [7], and their use in prediction, inference, and optimization was surveyed extensively by Peherstorfer et al. [8]. Subsequent studies extended the expressiveness of MF models to heterogeneous responses, multiple LF sources, and neural-network-based fusion [34–37], and LF information has also been used to improve the placement of the HF samples themselves [38].

To reduce the number of HF evaluations further, MF surrogates have been combined with sequential optimization. Huang et al. [10] proposed multi-fidelity sequential Kriging optimization, which selects the next design and its fidelity simultaneously by considering both the expected improvement (EI) and the evaluation cost. MF bandits and information-theoretic acquisition functions were developed in machine learning [39–41], and variable-fidelity EI and cost-aware MFBO were developed in engineering [11,13,42]. Although these studies differ in surrogate form and acquisition strategy, they all infer the LF–HF relation from numerical observations alone, namely LF and HF data. When HF observations are few and the LF model misses a local discrepancy structure, the correction between fidelities is therefore difficult to learn accurately [9,17]. This calls for information sources other than numerical data in MF models.

### 2.2 BO with LLMs

Studies that combine LLMs with BO can be distinguished by the component of BO into which the LLM output enters. LLAMBO [27], an early systematic approach, expresses the optimization problem and the observation history in natural language and uses the LLM throughout initialization, candidate generation, and surrogate modeling; its LLM-based surrogate directly predicts the objective value and its uncertainty at candidate points. In-context-learning-based BO has also been attempted for chemical reaction optimization [43]. Instead of objective values, LGBO [31] uses promising regions identified by the LLM, incorporates them into the GP mean as a region-lifted preference, and scales the shift by the confidence of the LLM judgment so that an incorrect judgment does not distort the surrogate excessively. LABO [28] treats numerical predictions from the LLM as an LF data source: the LLM takes the place of the LF model, a Kennedy–O'Hagan-type discrepancy model combines the LLM predictions with actual observations, and a gating criterion decides whether a candidate is evaluated by the LLM prediction or by an actual experiment. LILO [32] uses the LLM to interpret natural-language feedback into pairwise preference labels and learns a latent utility function with a pairwise GP; its ablation showed that pairwise comparisons provide more reliable estimates than directly requesting scalar utilities. This agrees with the observation that LLMs make more consistent judgments in pairwise comparisons than in absolute scoring [29,30].

Table 1 summarizes, for each method, the question posed to the LLM, its output, and where the output enters. In the existing methods, the question concerns the objective or utility, and the output enters the surrogate mean, candidate generation, LF data, or a utility function. None of these methods uses the LLM output together with a physics-based LF model; LABO replaces the LF model with the LLM. In addition, only LABO calibrates the LLM output against observations through a discrepancy model, whereas LGBO scales its influence by the confidence of the LLM judgment. The proposed method is distinguished in that it retains the physics-based LF model, asks about the error of the LF model, uses the output as a knowledge basis of the MF surrogate, and calibrates the contribution and sign of this basis with HF data.

**Table 1.** Role of the LLM in LLM-based BO methods.

| Method | Question posed to the LLM | LLM output | Where the output enters |
|---|---|---|---|
| LLAMBO [27] | What is the objective value of this design? | Value and uncertainty | Surrogate, candidate generation, and initialization |
| LGBO [31] | Which region is promising? | Promising region (qualitative) | GP mean shift scaled by confidence |
| LABO [28] | What is the objective value of this design? | Value | LF data source replacing the LF model |
| LILO [32] | Which outcome does the user prefer? | Pairwise preference (qualitative) | Latent utility function |
| **LLM-Assisted MFBO (this study)** | **Which of two designs requires the more positive residual correction of the LF model?** | **Pairwise comparison or abstention (qualitative)** | **Knowledge basis of the MF surrogate** |

Methodologically, this study builds on the preference GP [44], which learns a latent function from pairwise comparisons, and on active learning that selects comparison pairs by information gain [45]. Whereas the latent function in preference-based BO [46,47] and in LILO represents what is good, the latent function in this study represents where the LF model errs relatively more, and it becomes a basis that is calibrated by HF data within the MF surrogate.

## 3. Proposed method: LLM-Assisted MFBO

This section presents LLM-Assisted MFBO, which uses expert knowledge as an additional information source of the MF surrogate. Section 3.1 explains why conventional MF models struggle with design-dependent discrepancies and why expert knowledge and LLMs are suitable for complementing them. Section 3.2 presents the key concept of combining LF data with a knowledge basis. Section 3.3 details (1) comparison queries based on expert knowledge, (2) the construction of the knowledge basis, and (3) calibration with HF data. Section 3.4 describes the overall procedure that couples the proposed surrogate with BO.

### 3.1 Motivation

The basic strategy of conventional MF surrogate modeling is to approximate the expensive HF response by exploiting information from an inexpensive LF model. A representative two-level MF model is [6]

$$y_{\mathrm{H}}(\mathbf{x}) = \rho^{\mathrm{data}}\, y_{\mathrm{L}}^{\mathrm{data}}(\mathbf{x}) + \delta(\mathbf{x}), \tag{1}$$

where $\mathbf{x}$ is the vector of design variables, $y_{\mathrm{L}}^{\mathrm{data}}(\mathbf{x})$ is the LF response given by LF data (or by an LF surrogate trained on them), $\rho^{\mathrm{data}}$ is a global scale between the LF and HF responses, and $\delta(\mathbf{x})$ is the discrepancy that the scaled LF response cannot explain. Abundant LF information captures the global trend of the HF response, and the limited HF observations are used to learn the discrepancy.

The accuracy of a conventional MF model therefore depends on how accurately $\delta(\mathbf{x})$ can be inferred from limited HF observations. If the difference between LF and HF is mainly a global scale or a smooth bias, a few HF data suffice. This assumption, however, fails when the LF model simplifies or omits certain physical phenomena. The missing mechanisms produce nonlinear or local differences, and sometimes discrepancies that become pronounced only under specific operating conditions. In the one-dimensional example of Fig. 1, the LF model is a linear-elastic model in which stress is proportional to strain, and the HF model is an elastoplastic model that accounts for plastic deformation after yield. The two models nearly coincide at low loads (region A), but their difference grows rapidly beyond the yield load (region B). A conventional MF surrogate built from a few HF observations merely interpolates between them and cannot reproduce the continued growth of the discrepancy in the high-load region, where no observations are available; there, its estimate of the discrepancy reverts to the mean.

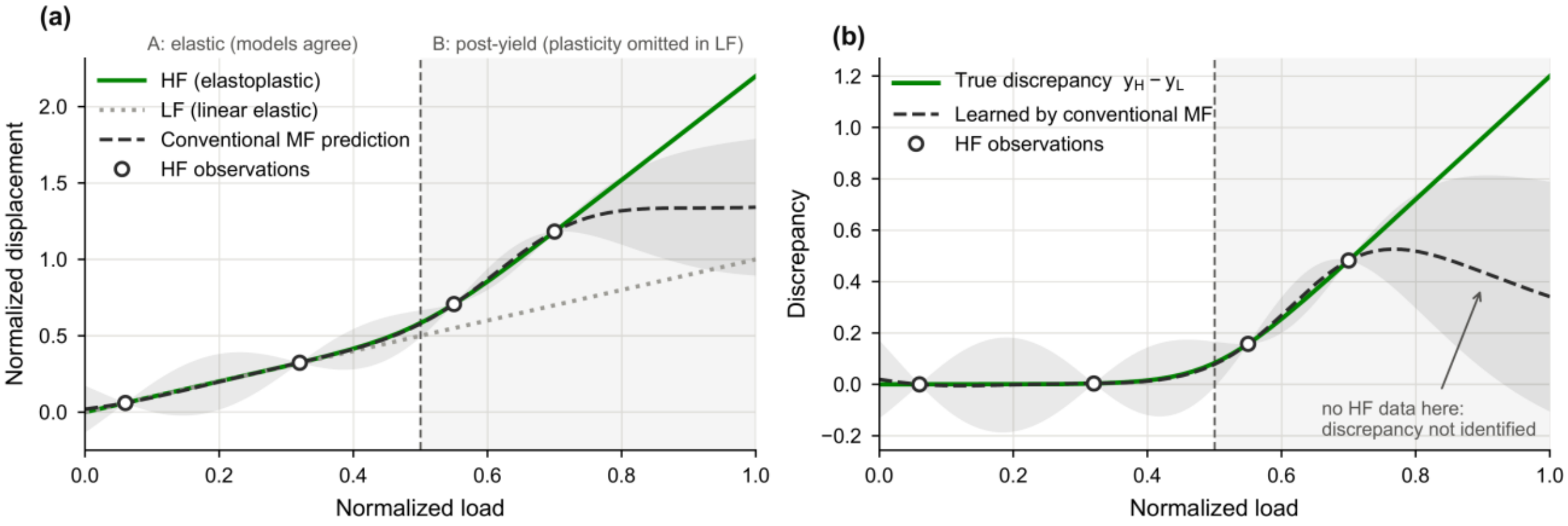

**Fig. 1.** One-dimensional example illustrating the limitation of conventional MF models (linear-elastic LF and elastoplastic HF; normalized load and displacement): (a) LF and HF responses, four HF observations, and prediction of the conventional MF surrogate (shading: ±2σ); and (b) true discrepancy and discrepancy learned by the conventional MF model (dashed line: yield load).

This limitation becomes clearer when the discrepancy is examined directly. After accounting for the global scale, the design-dependent correction is

$$d_{\rho}(\mathbf{x}) = y_{\mathrm{H}}(\mathbf{x}) - \rho^{\mathrm{data}}\, y_{\mathrm{L}}^{\mathrm{data}}(\mathbf{x}). \tag{2}$$

The correction $d_{\rho}(\mathbf{x})$ has a structure associated with the mechanisms missing from the LF model, but a few HF observations provide insufficient numerical evidence for that structure. A lack of numerical observations, however, does not imply a lack of knowledge about the cause of the discrepancy. In Fig. 1, an expert can judge that a larger correction is needed in B than in A without knowing the exact HF displacement. In general, experts know which physical mechanisms the LF model omits, which design variables affect those mechanisms, and under which conditions the difference between fidelities grows. Such knowledge cannot determine the values of $d_\rho(\mathbf{x})$, but it provides information on where the discrepancy grows or shrinks relative to other designs, that is, on the shape of the discrepancy over the design space.

**Why an LLM?** A direct way to incorporate qualitative knowledge is for the expert to encode it as equations. Actual knowledge, however, is a natural-language description in which several mechanisms are conditionally intertwined; which effect dominates between two designs with competing characteristics is not stated explicitly, and the weight of each mechanism is difficult to quantify. An LLM can interpret such a description and extend it to comparative judgments on two specific designs [48,49]. In the proposed method, the LLM output is (i) restricted to relative comparisons, (ii) allowed to abstain when the evidence is insufficient, and (iii) given a final contribution that is determined by HF data.

The central question of this study is therefore as follows. Instead of learning the entire discrepancy from a few HF observations, can expert knowledge of the physics omitted in the LF model be structured by an LLM and used as an additional information source of the MF model? And does its effect depend more on the capability of the LLM or on the quality of the knowledge?

## 3.2 Key concept

The key idea behind the proposed method is to retain the numerical information of LF data while complementing the structure of the discrepancy with expert knowledge. The physical differences between the LF and HF models described by the expert are interpreted by the LLM to build a separate knowledge basis, and the HF response is expressed as

$$y_{\mathrm{H}}(\mathbf{x}) = \rho^{\mathrm{data}}\, y_{\mathrm{L}}^{\mathrm{data}}(\mathbf{x}) + \rho^{\mathrm{LLM}}\, y_{\mathrm{L}}^{\mathrm{LLM}}(\mathbf{x}) + \tilde{\delta}(\mathbf{x}), \tag{3}$$

where the LF response $y_{\mathrm{L}}^{\mathrm{data}}(\mathbf{x})$ serves as the LF basis, $y_{\mathrm{L}}^{\mathrm{LLM}}(\mathbf{x})$ is the knowledge basis that represents the expected shape of the discrepancy, $\tilde{\delta}(\mathbf{x})$ is the residual discrepancy not explained by the two bases, and $\rho^{\mathrm{data}}$ and $\rho^{\mathrm{LLM}}$ determine the contributions of LF data and expert knowledge, respectively. Eq. (3) decomposes $\delta(\mathbf{x})$ in Eq. (1) into a part explained by knowledge and a remainder to be learned from data. When the knowledge basis explains the dominant shape of the discrepancy, the structure that the residual GP must learn from a few HF data is reduced.

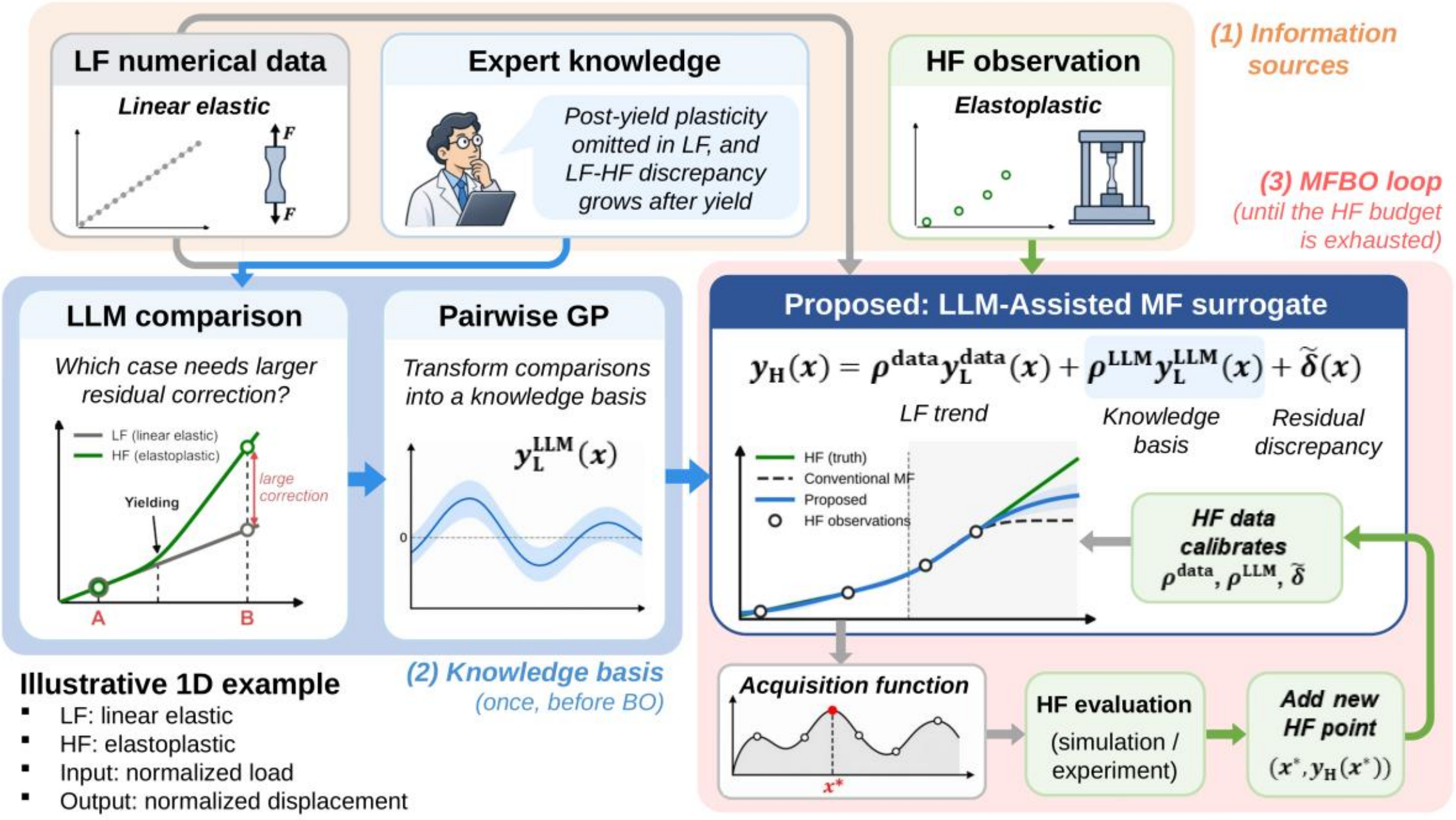

**Fig. 2.** Overview of LLM-Assisted MFBO, illustrated with the one-dimensional example of Fig. 1: (1) information sources, namely LF data, expert knowledge, and HF observations; (2) knowledge basis built once before BO, in which the LLM compares the residual corrections of actively selected design pairs (A, B, or TIE for abstention) and a pairwise GP converts these comparisons into a continuous knowledge basis $y_{\mathrm{L}}^{\mathrm{LLM}}(\mathbf{x})$; and (3) MFBO loop repeated until the HF budget is exhausted, in which HF data calibrate the LF basis, the knowledge basis, and the residual GP, the acquisition function selects the next HF evaluation, and the new HF point is added to the data.

As shown in Fig. 2, each information source plays a different role. LF data provide the global level and trend of the response, expert knowledge provides the shape of the discrepancy that the LF model cannot explain, and HF data provide the contributions of the two bases and the residual discrepancy. The LLM neither replaces the LF model nor predicts the HF response; its only role is to extend expert knowledge to comparative judgments on design pairs. This separation of roles leads to two properties. First, accurate knowledge allows the discrepancy structure to be exploited from the early stage, when HF data are scarce. Second, when no knowledge is available or the LLM abstains from all comparisons, $y_{\mathrm{L}}^{\mathrm{LLM}} \equiv 0$ and Eq. (3) becomes identical to Eq. (1).

## 3.3 Detailed strategies

This section describes the three components of the proposed method in the order in which they are executed.

### 3.3.1 Comparison queries based on expert knowledge

The information requested from the LLM is not the absolute HF response but the relative order of the corrections that the LF model cannot explain. Even an expert who knows the physical differences between the models can hardly state the HF response or the size of the discrepancy at a specific design as a number. The same expert, however, can judge qualitatively under which conditions an omitted phenomenon intensifies and whether a change in a design variable aggravates it. Building on this difference, the proposed method collects comparative judgments instead of numerical predictions.

**Query state and prompt.** Each design is presented to the LLM as a *state* $\mathbf{s}(\mathbf{x})$ that consists of the actual values of the design variables and the LF responses for each operating condition already computed at that design. The LF responses are included because the discrepancy can depend on the level of the LF response (e.g., when an omitted effect acts multiplicatively) and because the difference in LF responses between operating conditions is a cue to the strength of the omitted phenomenon. The same state is used as the input feature of the pairwise GP in Section 3.3.2, so the information observed by the LLM coincides with the space on which the knowledge basis is defined. Neither HF responses nor actual LF–HF errors are provided in any form. The system prompt consists of the problem definition (models, design variables, operating conditions, and objective), the expert knowledge (PHYSICS KNOWLEDGE), and the query rules (definition of the comparison target, prohibitions, and output format); the controlled study in Section 4 isolates the effect of knowledge quality mainly by changing PHYSICS KNOWLEDGE. The components of the query state and the system prompt are summarized in Tables A1 and A2, and the full text of the prompts is given in Appendix A.

**Comparison target.** For two designs $\mathbf{x}_i$ and $\mathbf{x}_j$, the comparison target is the signed residual correction in Eq. (2). The LLM is asked a question with the following meaning: "After a single global positive LF–HF scale has been accounted for, which design requires the more positive residual correction?" The LLM is not assumed to know or compute the value of the global scale. This condition excludes the scale difference common to all designs and focuses the judgment on the design-dependent discrepancy, which matches the division of roles in Eq. (3), where $\rho^{\mathrm{data}} y_{\mathrm{L}}^{\mathrm{data}}$ accounts for the global scale and $\rho^{\mathrm{LLM}} y_{\mathrm{L}}^{\mathrm{LLM}}$ for the design-dependent residual. When the LF model is defined as a special case of the HF model (e.g., a reduced model with some inputs fixed at nominal values), the reference scale can be set to one and $y_{\mathrm{H}} - y_{\mathrm{L}}^{\mathrm{data}}$ can be compared directly, as in the example of Fig. 1.

Note that the order of a comparison can depend on the global scale. Because the order of two designs reverses at $\rho = \Delta y_{\mathrm{H}} / \Delta y_{\mathrm{L}}^{\mathrm{data}}$, where $\Delta$ denotes the difference between the two designs, what the LLM compares is not the order at a particular value of $\rho$ but an order that is stable over a plausible range of global scales. The prompt instructs the LLM to answer TIE when no robust qualitative ordering can be justified; in the battery problem, all 13 A/B judgments made by the LLM agreed with the true order over the entire range $\rho \in [0.55, 1.0]$, whereas 4 of the 11 abstained pairs reversed their order within this range (Appendix E.4).

For each pair, the LLM returns one of A, B, and TIE together with a brief rationale in structured JSON. When the order cannot be justified on physical grounds alone, the LLM abstains with TIE; TIE thus denotes abstention due to insufficient evidence, not equality of the two residual corrections, and it is not converted into an arbitrary preference:

$$r_{ij}^{\mathrm{LLM}} = \begin{cases} +1, & \mathbf{x}_i \succ \mathbf{x}_j \text{ (A)} \\ -1, & \mathbf{x}_j \succ \mathbf{x}_i \text{ (B)} \\ 0, & \text{TIE} \end{cases} \tag{4}$$

where $\mathbf{x}_i \succ \mathbf{x}_j$ denotes that $\mathbf{x}_i$ requires the more positive residual correction. This label is not an inequality computed from HF data but an expected order based on expert knowledge. With the set of valid pairs $\mathcal{P} = \{(i, j, r_{ij}^{\mathrm{LLM}}) \mid r_{ij}^{\mathrm{LLM}} \in \{-1, +1\}\}$, TIE responses are excluded from the comparison likelihood. Box 1 shows examples of a judgment and an abstention in the battery problem of Section 4.

**Box 1. Examples of queries and responses (battery problem, P1; rationales paraphrased from the LLM output).** The design variables are scale factors relative to the reference values ($\lambda_L$: thickness, $\lambda_R$: particle radius) and porosities $\varepsilon$, and the LF responses are the energy densities (Wh/L) computed under the moderate and demanding conditions by the single-particle model (SPM), the LF model of Section 4.1. For brevity, only the design variables relevant to each judgment are shown.

**Judgment A.** Design A: $\lambda_{L_n}$ 0.98, $\lambda_{L_p}$ 1.13, $\lambda_{L_s}$ 1.01, $\varepsilon_n$ 0.28, $\varepsilon_p$ 0.37 (LF 883.8 / 775.3). Design B: $\lambda_{L_n}$ 1.13, $\lambda_{L_p}$ 0.93, $\lambda_{L_s}$ 0.86, $\varepsilon_n$ 0.22, $\varepsilon_p$ 0.29 (LF 1127.6 / 931.9). Rationale: "A has higher porosity in both electrodes and a thinner negative electrode, so the electrolyte transport limitation omitted by SPM is smaller. This advantage outweighs the penalty of its thicker positive electrode and separator; B therefore requires the more negative residual correction, and the correction for A is more positive." (True residual discrepancy: A +70.7, B −152.6 Wh/L.)

**Abstention (TIE).** Design A has a thicker negative electrode and separator and a slightly lower negative-electrode porosity ($\lambda_{L_n}$ 1.14, $\lambda_{L_s}$ 1.12, $\varepsilon_n$ 0.25), whereas design B has a thicker positive electrode and a lower positive-electrode porosity ($\lambda_{L_p}$ 1.14, $\varepsilon_p$ 0.34). Rationale: "A and B carry electrolyte transport burdens in opposite directions, and particle-size effects are partly represented in SPM, so the order of the remaining signed corrections cannot be determined with confidence."

The advantage of pairwise queries is that they provide information about the relative ordering of the discrepancy without requiring absolute magnitudes. If qualitative knowledge is used to answer how large a correction should be, the numerical scales and overconfidence that differ from model to model are passed directly to the surrogate. In contrast, which design requires the larger signed correction is a question that relative knowledge of the omitted physics can answer directly. Comparisons with other information formats, such as numerical predictions and scalar scores, are presented in Section 4.5.

### 3.3.2 Construction of the knowledge basis by a pairwise GP

The role of the pairwise GP is to convert a finite set of comparison labels into a continuous knowledge basis that can be evaluated over the entire design space. To this end, an unknown latent function $u(\mathbf{x})$ is introduced and learned so that differences between two points explain the comparison outcomes [44]. This latent function represents not the HF response but the relative size of the residual correction implied by expert knowledge. Its input is the state $\mathbf{s}(\mathbf{x}) = [\tilde{\mathbf{x}},\ \tilde{\mathbf{y}}_{\mathrm{L}}(\mathbf{x})]$, which combines the design variables normalized to $[0,1]$, $\tilde{\mathbf{x}}$, with the normalized LF responses for each operating condition, $\tilde{\mathbf{y}}_{\mathrm{L}}(\mathbf{x})$, so as to match the state observed by the LLM; at an arbitrary location in the continuous design space, $\tilde{\mathbf{y}}_{\mathrm{L}}(\mathbf{x})$ is computed with the LF surrogate:

$$u(\mathbf{x}) \sim \mathcal{GP}\left(0,\ k_{\mathrm{pref}}\big(\mathbf{s}(\mathbf{x}), \mathbf{s}(\mathbf{x}'); \boldsymbol{\theta}_{\mathrm{pref}}\big)\right), \qquad P\big(\mathbf{x}_i \succ \mathbf{x}_j \mid u\big) = \Phi\left(\frac{u(\mathbf{x}_i) - u\big(\mathbf{x}_j\big)}{\sqrt{2}\,\sigma_p}\right), \tag{5}$$

where $k_{\mathrm{pref}}$ is a radial basis function (RBF) kernel with hyperparameters $\boldsymbol{\theta}_{\mathrm{pref}}$, $\Phi$ is the standard normal cumulative distribution function, and $\sigma_p$ is the scale of stochastic variation in the judgments (fixed at $\sigma_p = 1$ for identifiability with the kernel amplitude). The posterior is computed by the Laplace approximation, and the kernel hyperparameters are estimated by maximizing the Laplace-approximated marginal likelihood. A single judgment supports $u(\mathbf{x}_i) > u\big(\mathbf{x}_j\big)$ without fixing the size of the difference, and the pairwise GP forms its posterior by accounting jointly for the relative relations shared by multiple comparisons and the correlation structure of the input space.

**Active pair querying.** Because the number of possible pairs grows quadratically with the number of candidates, it is important to select informative pairs under the same query budget $B$. The candidate pairs $\mathcal{E}$ are all pairs of anchors chosen by farthest-point sampling in the state space among designs with LF data, and the queries are made sequentially before BO. At each step, the pair that maximizes the expected reduction in uncertainty about the shape of the knowledge basis under the current posterior is selected. Let $\bar{I}_{ab}$ be the expected Fisher information of a probit observation on a candidate pair $(a, b)$ and $v_{ab}$ the posterior variance of the latent difference. The latent covariance after the observation then decreases by the rank-one update $\kappa_{ab}\mathbf{c}_{ab}\mathbf{c}_{ab}^{\mathrm{T}}$ with $\kappa_{ab} = \bar{I}_{ab}/(1 + \bar{I}_{ab}v_{ab})$, where $\mathbf{c}_{ab}$ is the difference in posterior covariance between anchors $a, b$ and the points of a reference set $\mathcal{R}$ spread over the design space (derivation in Appendix C.1). Because the knowledge basis is standardized and thus invariant to constant shifts, the criterion is the expected reduction of the centered variance over $\mathcal{R}$:

$$(a^\star, b^\star) = \arg\max_{(a,b)\in\mathcal{E}\setminus\mathcal{Q}} \kappa_{ab}\,\frac{1}{|\mathcal{R}|}\sum_{m\in\mathcal{R}} (\,c_{ab,m} - \bar{c}_{ab})^2, \tag{6}$$

where $c_{ab,m}$ is the element of $\mathbf{c}_{ab}$ at reference point $m$, $\bar{c}_{ab}$ is the mean of these elements, and $\mathcal{Q}$ is the set of pairs already queried. This criterion favors pairs whose judgment is uncertain and, at the same time, broadly influences the shape of the basis over the design space [45]. A TIE response does not change the posterior, but the pair is added to $\mathcal{Q}$. Comparisons with other pair-selection strategies are presented in Section 4.5.

**Knowledge basis.** The posterior mean $\mu_u(\mathbf{x})$ of the trained pairwise GP has an arbitrary latent scale set by the likelihood. It is therefore standardized to zero mean and unit variance over the reference set $\mathcal{R}$, and the result is used as the knowledge basis $y_{\mathrm{L}}^{\mathrm{LLM}}(\mathbf{x})$; the coefficient $\rho^{\mathrm{LLM}}$ learned subsequently determines its magnitude and sign. The knowledge basis is neither an HF prediction nor the size of the discrepancy; it is a shape that indicates where a relatively more positive or more negative correction is expected after the global scale has been accounted for. Fig. 3 illustrates this process for the example of Fig. 1, together with the proposed surrogate (Section 3.3.3) built from the same HF observations. Because the knowledge basis provides the shape of the post-yield discrepancy, the proposed surrogate follows the HF response even in the high-load region without HF observations.

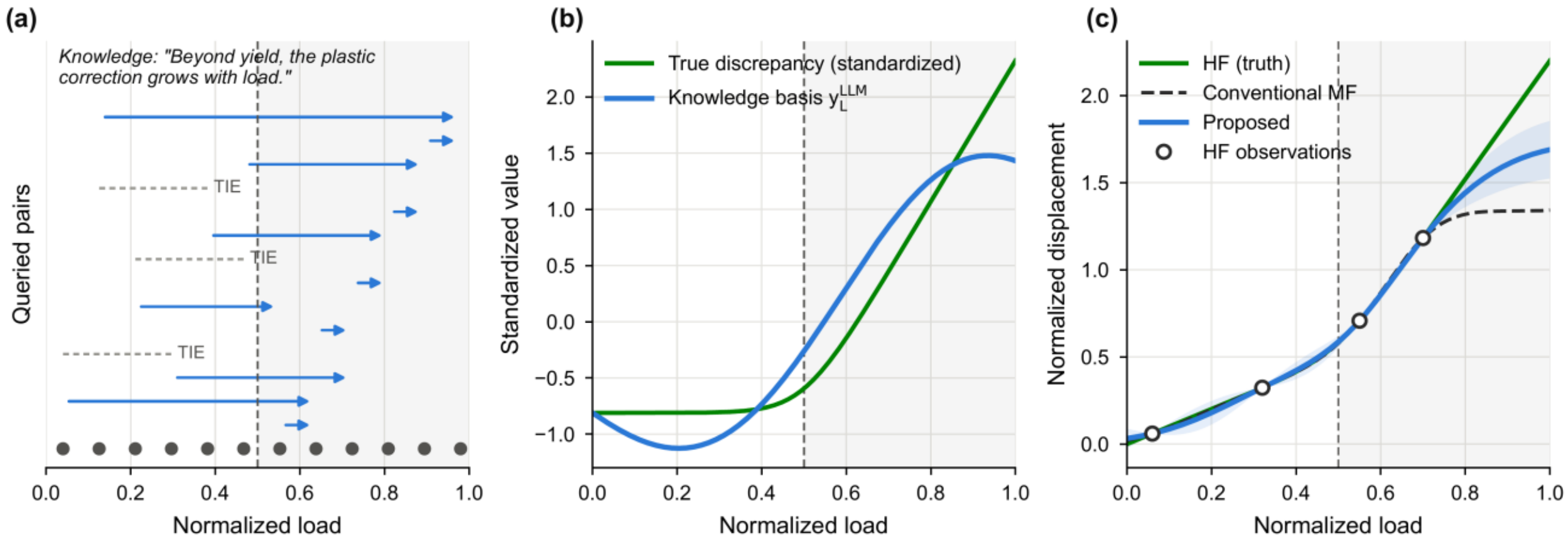

**Fig. 3.** Conversion of pairwise comparisons into a continuous knowledge basis for the example of Fig. 1: (a) pairwise judgments based on the single statement "beyond yield, the plastic correction grows with load" (arrows: design requiring the larger correction; dashed lines: TIE; judgments that apply the statement directly are used in place of the LLM in this conceptual example); (b) knowledge basis obtained by standardizing the posterior mean of the pairwise GP, and standardized true discrepancy (not used for training); and (c) proposed surrogate and conventional MF surrogate built from the same four HF observations (shading: ±2σ of the proposed surrogate).

### 3.3.3 Calibration with HF data

HF observations calibrate the magnitude and sign of the knowledge contribution, while its shape is determined by the pairwise GP. In Eq. (3), the LF basis $y_{\mathrm{L}}^{\mathrm{data}}(\mathbf{x})$ is evaluated with the LF surrogate trained on LF data, which provides the LF basis at any design, including HF observation points and candidates. The residual discrepancy $\tilde{\delta}(\mathbf{x})$ is modeled as a zero-mean GP with a Matérn-5/2 kernel with automatic relevance determination (ARD) on the normalized design variables. The coefficients are constrained as $\rho^{\mathrm{data}} \geq 0$ and $\rho^{\mathrm{LLM}} \in \mathbb{R}$ so that HF data can re-evaluate the sign of the knowledge basis. The coefficients and the kernel hyperparameters are estimated jointly by maximizing the marginal likelihood of the HF data, with the coefficients profiled out by constrained generalized least squares (GLS); the resulting predictive mean $\mu_{\mathrm{H}}(\mathbf{x})$ and standard deviation (SD) $\sigma_{\mathrm{H}}(\mathbf{x})$ take the standard form of GP regression with an explicit mean basis (Appendix D.1). The predictive variance is that of the residual GP conditional on the two bases and the estimated coefficients; the uncertainties of the pairwise GP and of the coefficient estimates are not propagated.

**Interpretation of the coefficients.** When the knowledge basis agrees with the HF residual response, HF data assign a large positive $\rho^{\mathrm{LLM}}$; when the basis has no explanatory power, the coefficient stays near zero; and when the basis opposes the actual trend, the coefficient becomes negative. Two structural properties follow from this construction (Appendix C.2). First, if all responses are TIE, there are no valid comparisons, the posterior mean of the pairwise GP is constant, and the knowledge basis vanishes ($y_{\mathrm{L}}^{\mathrm{LLM}} \equiv 0$). The proposed surrogate then coincides with the conventional MF model in Eq. (1), so that, for the same initial design and acquisition optimization, the proposed method selects the same designs as conventional MFBO. Second, reversing every valid comparison yields the basis $-y_{\mathrm{L}}^{\mathrm{LLM}}$; because $\rho^{\mathrm{LLM}}$ is not sign-constrained, HF calibration flips the sign of the coefficient and yields the same prediction. The proposed method thus returns to the conventional method when no knowledge is available, and HF data correct the LLM even when its judgments are globally reversed. In contrast, knowledge whose shape is only partially correct cannot be corrected by the sign alone, so the quality of knowledge is evaluated experimentally in Sections 4.3 and 4.4.

### 3.4 Overall procedure

The overall procedure of the proposed method consists of the following steps, as illustrated in Fig. 2 and detailed in Algorithm 1 (Appendix D.3). Steps 1–3 are performed once before BO, and Steps 4–6 are repeated until the HF budget is exhausted.

- **Step 1 (problem and knowledge definition):** Define the LF/HF models, design variables, operating conditions, and objective; the expert describes the mechanisms omitted in the LF model and the conditions under which their influence grows (PHYSICS KNOWLEDGE in the prompt).
- **Step 2 (LF information):** Generate LF data with the LF model and train the LF surrogate for use in the continuous design space. Construct anchors and candidate pairs in the state space from designs with LF data.
- **Step 3 (knowledge basis):** Sequentially select $B$ pairs by active pair querying (Eq. (6)) and query the LLM (Section 3.3.1); train the pairwise GP and standardize its posterior mean to construct the knowledge basis $y_{\mathrm{L}}^{\mathrm{LLM}}$ (Section 3.3.2). If all responses are TIE, $y_{\mathrm{L}}^{\mathrm{LLM}} \equiv 0$.
- **Step 4 (surrogate calibration):** Calibrate $\rho^{\mathrm{data}}$, $\rho^{\mathrm{LLM}}$, and the residual-GP hyperparameters jointly with the current HF data (Section 3.3.3).
- **Step 5 (next design):** Compute the EI [3] from the predictive mean and SD of the proposed surrogate and maximize it to select the next HF evaluation (Appendix D.2).
- **Step 6 (HF evaluation and update):** Evaluate the selected design with the HF model, add it to the data, and return to Step 4 if HF budget remains.

In this structure, knowledge extraction and optimization are clearly separated. The LLM is queried only $B$ times before BO (24 times in most experiments of this study), and only the coefficients and the residual GP are updated with HF observations during BO. Because HF data are never fed back to the LLM, there is no path through which the LLM could access HF information. In summary, LF data provide the baseline response, expert knowledge provides the relative shape of the discrepancy, and HF data determine the contributions of the two sources. Section 4 examines whether this separation of roles works in an actual engineering problem and what its effect depends on.

## 4. Numerical study

This section validates the proposed method on a seven-dimensional lithium-ion battery design problem in which the Doyle–Fuller–Newman (DFN) model and the SPM, both implemented in PyBaMM [50], serve as the HF and LF models, respectively. In this problem, the physics omitted in the LF model is conditionally intertwined with the operating conditions and design variables, and the design-dependent discrepancy remaining after the global scale determines the ranking of the HF responses. The experiments were designed to answer the following research questions (RQs), which correspond to the contributions stated in Section 1.

- **RQ1 (knowledge as an information source):** Does the knowledge basis accelerate MFBO, and what information does it provide compared with LF data? (Section 4.2)
- **RQ2 (knowledge vs. LLM):** Does performance depend more on the quality of the expert knowledge or on the choice of LLM? (Section 4.3)
- **RQ3 (contribution of each knowledge statement):** How much does each knowledge statement contribute to performance, and how can this information be used in design? (Section 4.4)

- **RQ4 (lack of knowledge):** Does the proposed method reduce to conventional MFBO when no knowledge is available? (Section 4.3)

In addition, Section 4.5 briefly examines the design choices regarding the target and format of the questions posed to the LLM and the pair-selection strategy.

**Methods and metrics.** The baseline methods are (i) HF-only BO, which uses a GP with a constant mean and EI; (ii) conventional MFBO, which uses the MF surrogate in Eq. (1); and (iii) the proposed method, which uses the surrogate in Eq. (3). The three methods share the same HF surrogate implementation (ARD Matérn-5/2 kernel, profile GLS, and multi-start L-BFGS-B; Appendix Table B1), the same EI optimization, and the same initial HF designs; they differ only in the mean basis. Performance is summarized by the area under the curve (AUC), obtained by integrating the best-observed objective value up to each HF evaluation after the initial design with the trapezoidal rule and dividing by the length of the interval; its seed-to-seed SD is reported as a measure of variability. The regret AUC, normalized by the maximum and minimum of all HF evaluations in an experiment ($r_n = \left(y^{\max} - y_n^{\text{best}}\right)/\left(y^{\max} - y^{\min}\right)$; lower is better), is also reported. In addition, the Spearman rank correlation between the knowledge basis and the true residual discrepancy $d^{\star} = y_{\text{H}} - \rho^{\star} y_{\text{L}}$ (basis accuracy) is computed post hoc on 256 designs for which both LF and HF responses are available. None of these quantities is used during BO. All comparisons are paired by seed, with each seed sharing the same initial HF design; 95% confidence intervals (CIs) are obtained by paired bootstrap, and significance is assessed by the Wilcoxon signed-rank test [51] with Holm correction [52]. The LLM query plan is generated once per condition and shared by all seeds.

## 4.1 Problem definition and expert knowledge

The design problem maximizes the volumetric discharge energy density under two operating conditions and has seven design variables; it is based on the Chen2020 parameter set [53] for the LG M50 (21700) cell. The HF model is the DFN model [54], which resolves the through-cell distributions of electrolyte concentration and potential, and the LF model is the SPM [55], which omits electrolyte dynamics and approximates each electrode by a single particle. Both models were implemented in PyBaMM under isothermal conditions with the IDAKLU solver (relative tolerance $10^{-4}$, absolute tolerance $10^{-6}$) and an output interval of 30 s. Each design is discharged at constant current to the lower voltage limit (2.5 V) under a moderate condition (5 A, 25 °C) and a demanding condition (15 A, 15 °C). The objective is the average over the two conditions of the discharge energy divided by the volume of the electrochemical stack:

$$\max_{\mathbf{x}} y(\mathbf{x}) = \frac{1}{2} \sum_{s \in \{\text{mod}, \text{dem}\}} \frac{E_s(\mathbf{x})}{A\left(L_n(\mathbf{x}) + L_s(\mathbf{x}) + L_p(\mathbf{x})\right)} \quad [\text{Wh/L}]. \tag{7}$$

where $E_s$ is the discharge energy under condition $s$ (moderate or demanding), $A$ is the electrode area, and $L_n$, $L_p$, and $L_s$ are the thicknesses of the negative electrode, positive electrode, and separator, respectively. The seven design variables are the thickness scale factors of the negative electrode, positive electrode, and separator, the porosities of the two electrodes, and the particle-radius scale factors of the two electrodes (ranges in Appendix Table B2). Because the active-material fraction decreases as the porosity increases, a thin and porous electrode relieves transport limitations but stores less energy, which creates a trade-off.

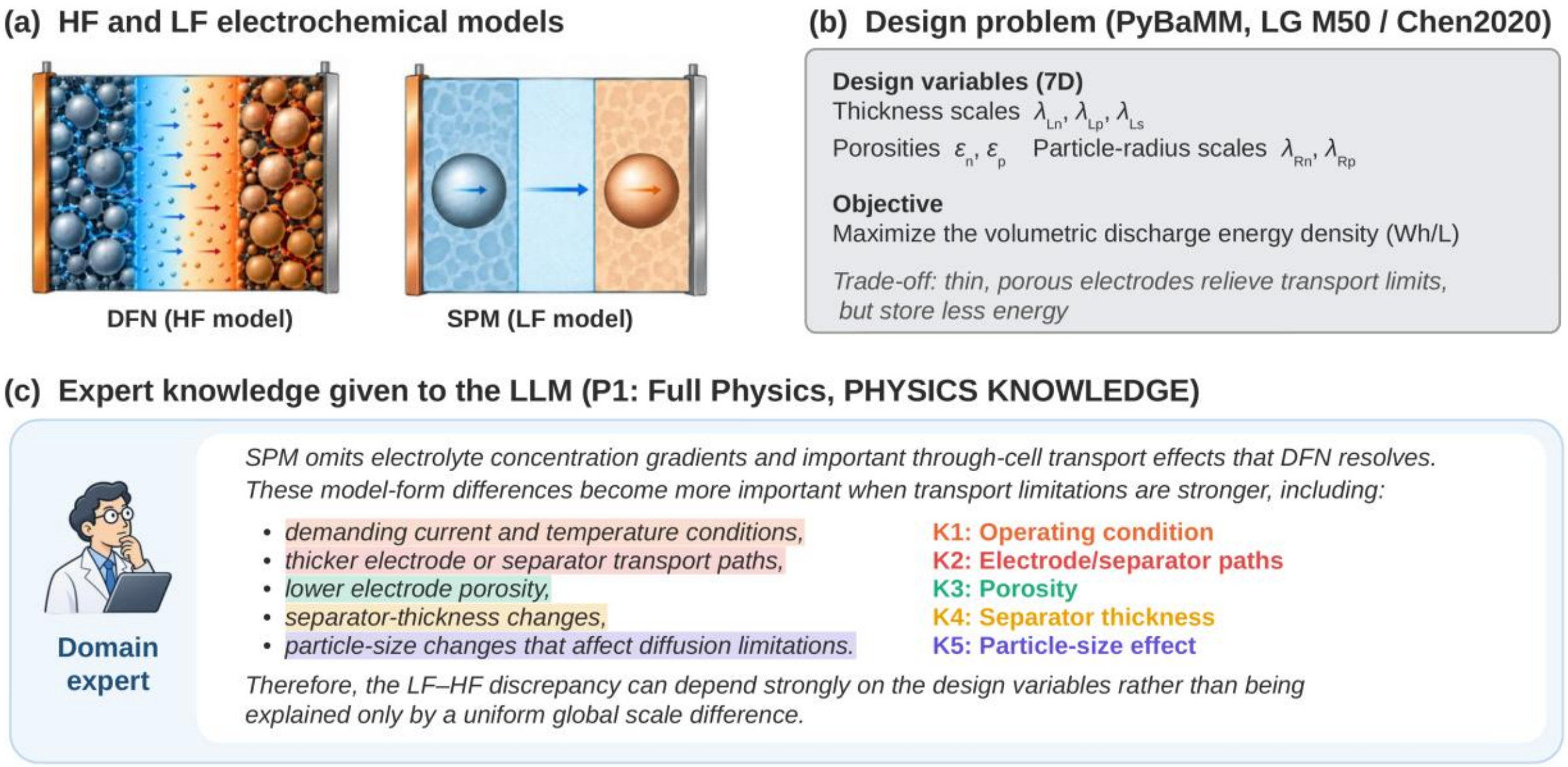

**Fig. 4.** Lithium-ion battery design problem: (a) physical differences between the HF (DFN) and LF (SPM) models; (b) design variables and objective; and (c) PHYSICS KNOWLEDGE of P1 (Full Physics) provided to the LLM, with the five conditions labeled K1–K5.

This problem is a typical case in which the design-dependent discrepancy dominates, and it is particularly challenging for conventional MFBO. Over the 256 designs for which both LF and HF responses are available, the Pearson correlation between the SPM and DFN objectives is $-0.019$ and the Spearman rank correlation is $-0.090$, which means that they are virtually uncorrelated, and the DFN response is lower than the SPM response at every design (least-squares estimate of the global positive scale, $\rho^{\star} = 0.766$). The LF model thus provides information on the overall response level but little information on the HF ranking, which is determined by the design-dependent discrepancy remaining after the global scale. In contrast, electrochemical knowledge of where this discrepancy grows is well known to experts.

The baseline knowledge provided by the expert (P1) consists of a statement distinguishing the models ("SPM omits electrolyte concentration gradients and important through-cell transport effects that DFN resolves"), five conditions under which these differences become more important (K1–K5, Fig. 4(c)), and a concluding statement that "the LF-HF discrepancy can depend strongly on the design variables rather than being explained only by a uniform global scale difference" (Appendix A.1). The query state consists of the seven design variables, the LF responses under the two conditions, and the scalar LF objective. The LF basis is a GP surrogate trained on the SPM results for 256 designs, and the designs proposed by BO were evaluated with actual DFN simulations. The settings of each experiment are summarized in Appendix Table B3.

## 4.2 Information-source ablation: contributions of LF data and expert knowledge

**Setup.** To separate the contributions of LF data and expert knowledge, four methods that differ in the information bases included in their mean functions were compared: (i) HF only, standard BO with a constant mean; (ii) HF+LF, conventional MFBO; (iii) HF+LLM, a model that uses only the knowledge basis without the LF basis (no intercept, $\rho^{\mathrm{LLM}} \geq 0$); and (iv) HF+LF+LLM, the proposed method. In HF+LLM, the LF responses are still part of the state observed by the LLM and of the input to the pairwise GP, but the LF values are not used as a basis. The two LLM-based methods share the same 24 query results (6 A, 7 B, and 11 TIE), so only 13 valid comparisons are available. For each of 20 seeds, the four methods start from the same seven initial DFN evaluations and perform eight sequential DFN evaluations (80 BO trajectories in total).

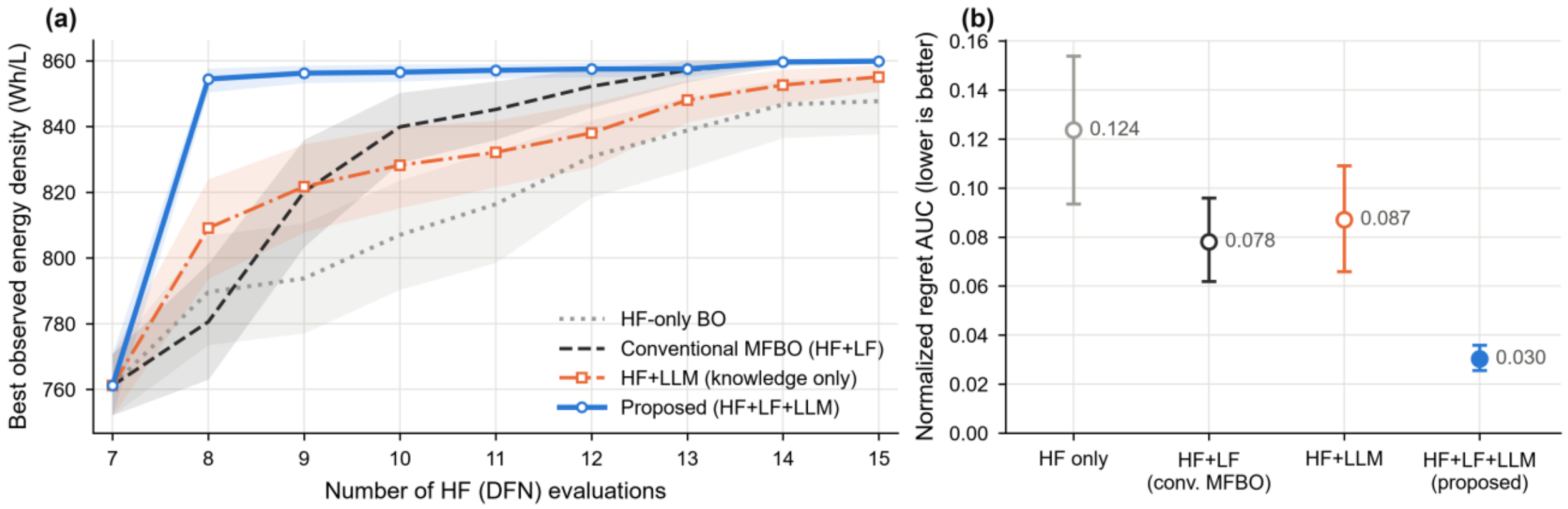

**Fig. 5.** Information-source ablation (battery problem, 20 seeds): (a) mean best-observed DFN energy density versus the number of HF evaluations (shading: 95% bootstrap CI); and (b) mean normalized regret AUC with 95% CIs (lower is better).

**Results.** In Fig. 5 and Table 2, the proposed method converged much faster than the other methods, particularly during the early BO iterations. It reached 854.5 Wh/L on average immediately after the first BO iteration, achieving most of the improvement over the initial best value (761 Wh/L) with a single HF evaluation. At the same point, HF+LF, HF+LLM, and HF only remained at about 780, 809, and 790 Wh/L, respectively. The regret AUC of the proposed method was 0.0303, which is 61.2% lower than that of conventional MFBO (0.0781), 75.5% lower than that of HF only (0.1237), and 65.2% lower than that of HF+LLM (0.0871). In seed-wise comparisons, the proposed method outperformed HF only in 19 of 20 seeds and HF+LF and HF+LLM in 18 seeds each, and the 95% CIs of the AUC differences excluded zero in all three comparisons (Appendix Fig. E1(b)). The SD of the AUC for the proposed method (0.0122) was about one-third of that for conventional MFBO (0.0396).

**Table 2.** Results of the information-source ablation (20 seeds; brackets denote 95% bootstrap CIs).

| Method | Regret AUC ↓ [95% CI] | AUC SD | Final best DFN [Wh/L] | Paired AUC gain of the proposed method [95% CI] | Seeds won |
|---|---|---|---|---|---|
| HF only | 0.1237 [0.0959, 0.1546] | 0.0698 | 847.70 | +0.0934 [0.0662, 0.1236] | 19/20 |
| HF + LF (conventional MFBO) | 0.0781 [0.0617, 0.0954] | 0.0396 | 859.96 | +0.0478 [0.0314, 0.0655] | 18/20 |
| HF + LLM (knowledge only) | 0.0871 [0.0663, 0.1097] | 0.0510 | 855.07 | +0.0568 [0.0356, 0.0801] | 18/20 |
| **HF + LF + LLM (proposed)** | **0.0303 [0.0256, 0.0362]** | **0.0122** | **859.87** | – | – |

**Complementarity.** Relative to HF only, the AUC decreased by 0.0456 when LF data were added, by 0.0366 when knowledge was added, and by 0.0934 when both were used. The seed-wise interaction (combined gain minus the two individual gains) was +0.011 (95% CI [−0.020, 0.041]) and did not differ significantly from zero, so the combined improvement was consistent with additive contributions from the two sources rather than with one cancelling the other. With both sources, the method outperformed HF only in 19 of 20 seeds (LF alone: 13/20; knowledge alone: 15/20). This agrees with the separation of roles in Section 3.2: the LF basis provides the level and global trend of the response but cannot explain the design-dependent discrepancy, whereas the knowledge basis provides the shape of the discrepancy but carries no information on the response level.

The proposed method and conventional MFBO reach the same optimal region (about 860 Wh/L) by the end of the HF budget, so the advantage of the proposed method lies in the number of HF evaluations needed to reach that level. With a target of 850 Wh/L, 80% of the seeds of the proposed method reached the target immediately after the first BO iteration, whereas the other three methods reached it in only 10–15% of the seeds, and conventional MFBO required 13 HF evaluations to reach 90% (Fig. 6(a)).

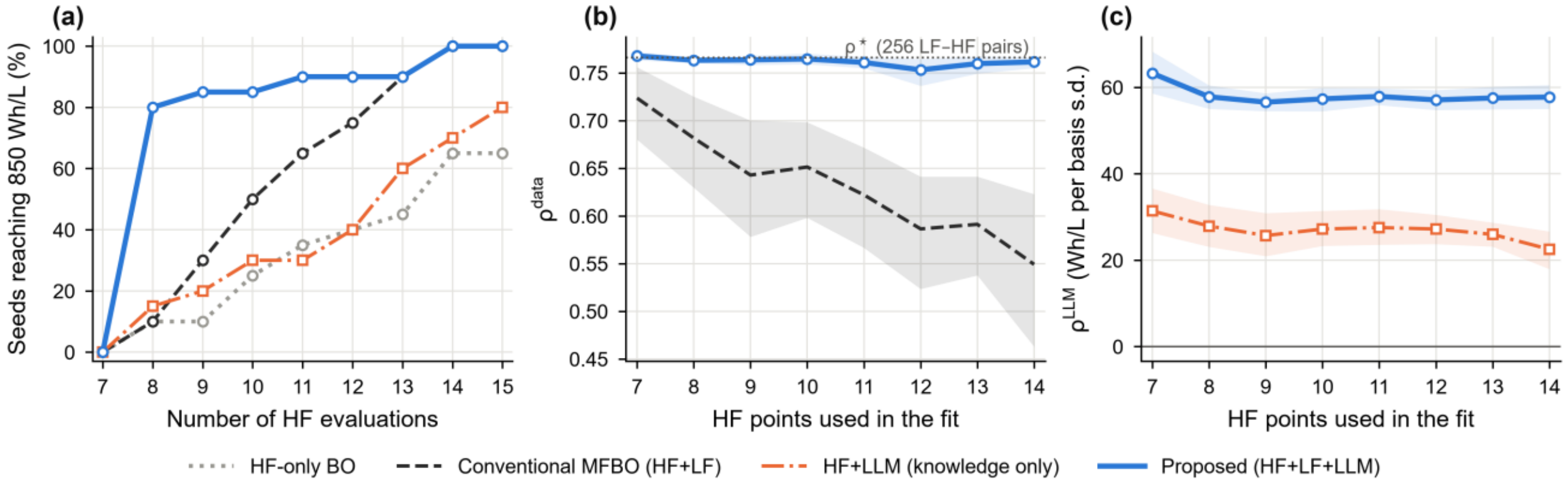

**Fig. 6.** Mechanism of the information-source ablation (battery problem, 20 seeds): (a) fraction of seeds reaching the 850 Wh/L target; (b) mean $\rho^{\mathrm{data}}$ calibrated with HF data (dashed line: global scale $\rho^{\star} = 0.766$ computed from 256 LF–HF pairs); and (c) mean $\rho^{\mathrm{LLM}}$ calibrated with HF data (shading: 95% bootstrap CI).

**Mechanism.** In Fig. 6(c), $\rho^{\mathrm{LLM}}$ of the proposed method was estimated as positive (about 63 Wh/L) from only the seven initial HF data and then stabilized at about 57 Wh/L. In other words, HF data supported the direction of the discrepancy shape suggested by the expert knowledge. More notable is the behavior of the LF coefficient. In the proposed method, $\rho^{\mathrm{data}}$ remained at about 0.76 (Fig. 6(b)) throughout the iterations, nearly identical to the global scale $\rho^{\star} = 0.766$ computed post hoc from 256 LF–HF pairs. In conventional MFBO, by contrast, $\rho^{\mathrm{data}}$ decreased from 0.72 to 0.55. This is interpreted as *misattribution*, in which the unexplained design-dependent discrepancy is absorbed into the global scale coefficient. The knowledge basis therefore not only explains the shape of the discrepancy directly but also keeps the LF coefficient at a physically meaningful value, which suggests that it reduces confounding between the global LF scale and the residual discrepancy when samples are few. Indeed, the knowledge basis built from only 13 valid comparisons reached a Spearman rank correlation of 0.96 with the true residual discrepancy $d^{\star}$ computed post hoc on 256 designs, and 0.87 even among the top 25% of designs by objective value. Without access to any HF values, the knowledge basis derived by the LLM from expert knowledge captured the relative ordering of the true residual discrepancy with high accuracy. The best designs found by all methods converged toward thin electrodes and separators, high positive-electrode porosity, and small particles (Appendix Fig. E1(a)), which agrees with the interpretation that relieving transport limitations under the demanding condition governs the volumetric energy density, as well as with the directions indicated by K1–K5.

## 4.3 Controlled study: knowledge quality or choice of LLM?

**Setup.** To determine whether the effect observed in the information-source ablation depends more on the quality of the expert knowledge or on the choice of LLM, a 4 × 4 factorial controlled study that varies the two factors independently was conducted. The protocol is the same as in Section 4.2 (the same 20 initial designs, HF 7→15, 24 active pair queries per condition, medium reasoning effort), and the four knowledge conditions differ primarily in the PHYSICS KNOWLEDGE component (Appendix A.1–A.4); for P3, domain-specific information is additionally anonymized to establish a strict no-knowledge control.

- **P1 Full Physics:** accurate and sufficient knowledge (including K1–K5).
- **P2 Partial Physics:** accurate but coarse partial knowledge. It states only that SPM simplifies electrolyte and through-cell transport, that the demanding scenario is more strongly transport-limited than the moderate scenario, and that “the relative change between the two supplied LF scenario responses may be used as a COARSE QUALITATIVE indicator of how strongly a design is affected by transport limitations"; it explicitly supplies no variable-specific rules.

- **P3 Minimal / No Physics:** no knowledge. The domain, the meaning of the variables, the model names, and the meaning of the operating conditions are anonymized in both the system prompt and the states ($x_1, \dots, x_7$, LF response 1 and 2).
- **P4 Misleading Physics:** deliberately reversed knowledge. The incorrect statements that "The LF SPM represents electrolyte concentration gradients and through-cell transport limitations more strongly than the HF DFN model" and that the discrepancy "is expected to become more important under weaker transport-limitation conditions" are presented as the authoritative prior for the comparison.

Four LLMs with reasoning capability were used (GPT-6 Astra, GPT-6.1 Sol, GPT-6 Sol, and GPT-5.6 Sol). Because an independent 24-query plan was generated for each of the 16 conditions (384 queries in total), the selected pairs themselves also depend on the knowledge and on the LLM responses. The 340 BO trajectories, including conventional MFBO (5,100 DFN evaluations), share the same initial designs, and the normalization uses the maximum and minimum of the DFN evaluations over all 17 methods (862.63 and 556.28 Wh/L). The P1 × GPT-6.1 Sol condition therefore follows the same trajectories as the proposed method in Section 4.2, but its AUC is reported as 0.0374 because of the different normalization. In this study, $\rho^{\mathrm{LLM}}$ was estimated without a sign constraint.

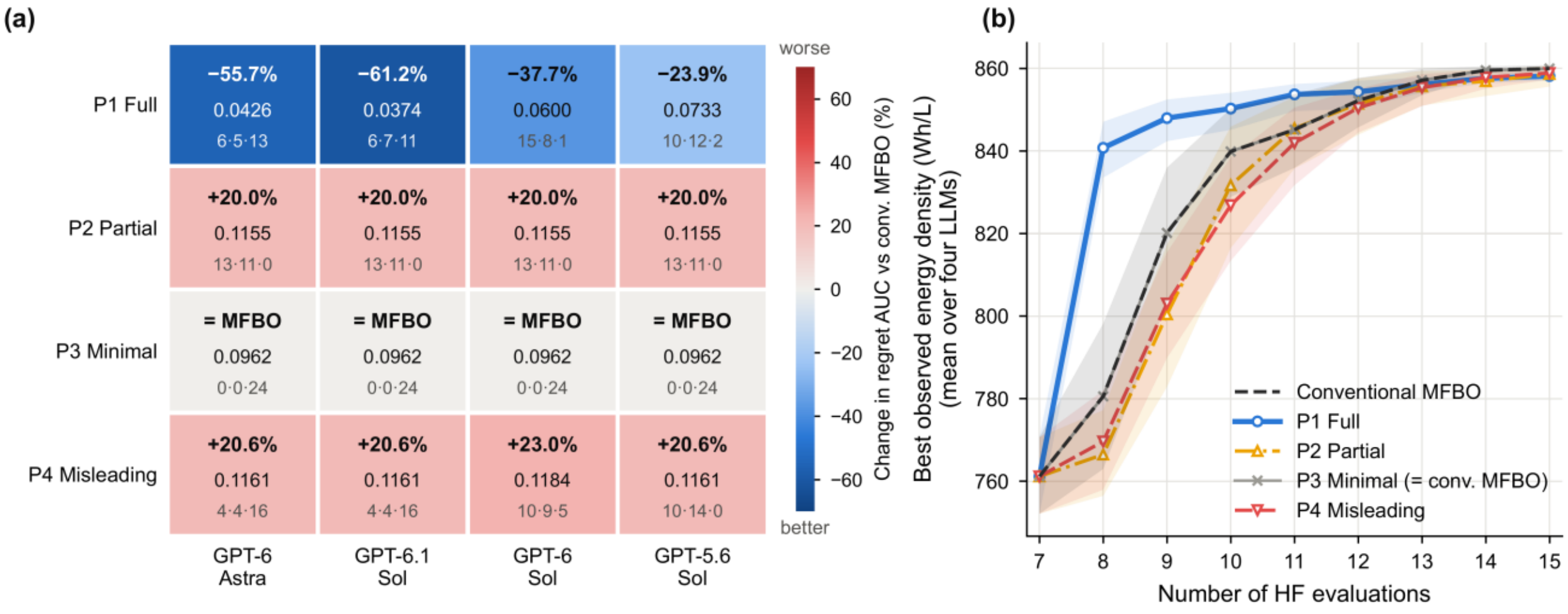

**Fig. 7.** Knowledge quality × LLM controlled study (battery problem, 20 seeds): (a) change in normalized regret AUC relative to conventional MFBO for the 16 conditions (blue: improvement; red: degradation; gray: identical; cell values from top to bottom: change, normalized regret AUC with conventional MFBO at 0.0962, and numbers of A, B, and TIE responses over the 24 queried pairs); and (b) mean convergence of the best-observed DFN objective for each knowledge condition, averaged over the four LLMs (shading: 95% bootstrap CI; black dashed line: conventional MFBO).

**Knowledge quality determines the direction of performance.** In Fig. 7(a), accurate knowledge (P1) improved the mean performance over conventional MFBO with all four LLMs. The regret AUC decreased by 61.2% with GPT-6.1 Sol, 55.7% with GPT-6 Astra, 37.7% with GPT-6 Sol, and 23.9% with GPT-5.6 Sol (44.6% on average), and the proposed method outperformed conventional MFBO in 18, 17, 17, and 13 seeds, respectively. In Wilcoxon tests with Holm correction over all 16 conditions, the improvements with GPT-6.1 Sol ($p_{\mathrm{Holm}}$ = 0.0002), GPT-6 Astra (0.002), and GPT-6 Sol (0.024) were significant. The distributions of the seed-wise gains are shown in Appendix Fig. E2(a). The rate of reaching the 850 Wh/L target immediately after the first BO iteration was also far higher than that of conventional MFBO (10%): 80% with GPT-6.1 Sol, 65% with GPT-6 Astra, 45% with GPT-6 Sol, and 40% with GPT-5.6 Sol (Appendix Fig. E2(b)). In contrast, with partial knowledge (P2) and misleading knowledge (P4), the mean AUC was about 20% higher than that of conventional MFBO for all four LLMs, and the method underperformed in 14–16 seeds. Averaged over the four LLMs, P1 converged faster than conventional MFBO, whereas P2 and P4 converged more slowly (Fig.

7(b)). The same LLM thus changed the direction of performance depending on the knowledge. The size of this degradation, however, was less than half of the improvement under P1, and only one condition, P4 × GPT-5.6 Sol, remained significant after multiple-comparison correction. This is consistent with HF calibration attenuating the degradation even when the knowledge was inaccurate. Unlike a global reversal of the pairwise labels (Section 3.3.3), misleading knowledge generates new judgments and can alter the selected pairs, the abstentions, and the shape of the learned basis, which cannot be corrected by changing the coefficient sign alone.

**Changing the knowledge matters far more than changing the LLM.** A two-factor variance decomposition of the mean AUC over the 16 conditions showed that the knowledge condition explained 92.8% of the between-condition variance and the choice of LLM 1.9% (interaction: 5.3%). The mean absolute change in AUC when the knowledge was changed with the LLM fixed (0.0349) was about 6.3 times that when the LLM was changed with the knowledge fixed (0.0055). For example, changing the knowledge from P1 to P4 with GPT-6.1 Sol increased the AUC by a factor of 3.1, whereas the difference between the best and worst LLMs under P1 was only 0.036. This decomposition is based on condition means, each obtained from a single query plan generated for that condition. The conclusion was unchanged when the computation was performed per seed rather than on condition means: in each seed, the mean absolute change in AUC when the knowledge was changed with the LLM fixed (0.047) was about 3.2 times that when the LLM was changed with the knowledge fixed (0.015), and the effect of knowledge was larger in all 20 seeds (paired Wilcoxon $p < 10^{-6}$).

**Without knowledge, the LLM abstains and the method reduces to conventional MFBO.** Under the anonymized P3, all four LLMs abstained on all 96 queries. The knowledge basis consequently vanished, and the proposed method reduced to exactly the same trajectories and performance as conventional MFBO (AUC 0.0962) with every LLM (Section 3.3.3). Under misleading knowledge (P4), some LLMs abstained more often (16 TIE responses each for GPT-6 Astra and GPT-6.1 Sol), and HF calibration assigned small coefficients to the P4 basis. Without knowledge, the proposed method therefore returns to the conventional method; with knowledge, its quality determines the direction of performance. The tendency of LLMs that abstained more often under P1 to produce better bases, and the diagnostics for partial knowledge (P2), are reported in Appendix E.2.

## 4.4 Knowledge-level ablation: contribution of each knowledge statement

**Setup.** To quantify the contribution of each knowledge statement in P1, a leave-one-knowledge-out (LOKO) ablation was conducted. Five conditions in which exactly one line among K1–K5 was deleted from P1 (P1−K1, …, P1−K5) were compared with P1, with the rest of the prompt kept identical character by character. For each knowledge condition, 24 new active pair queries were made with GPT-6.1 Sol (high reasoning effort), so the queried pairs themselves also depend on the knowledge. The conditional importance of knowledge statement $k$ is defined as the seed-wise difference $\Delta\mathrm{AUC}_k = \mathrm{AUC(P1)} - \mathrm{AUC}(\mathrm{P1} - \mathrm{K}k)$ (best-observed AUC in Wh/L; a positive value indicates degradation when the statement is removed).

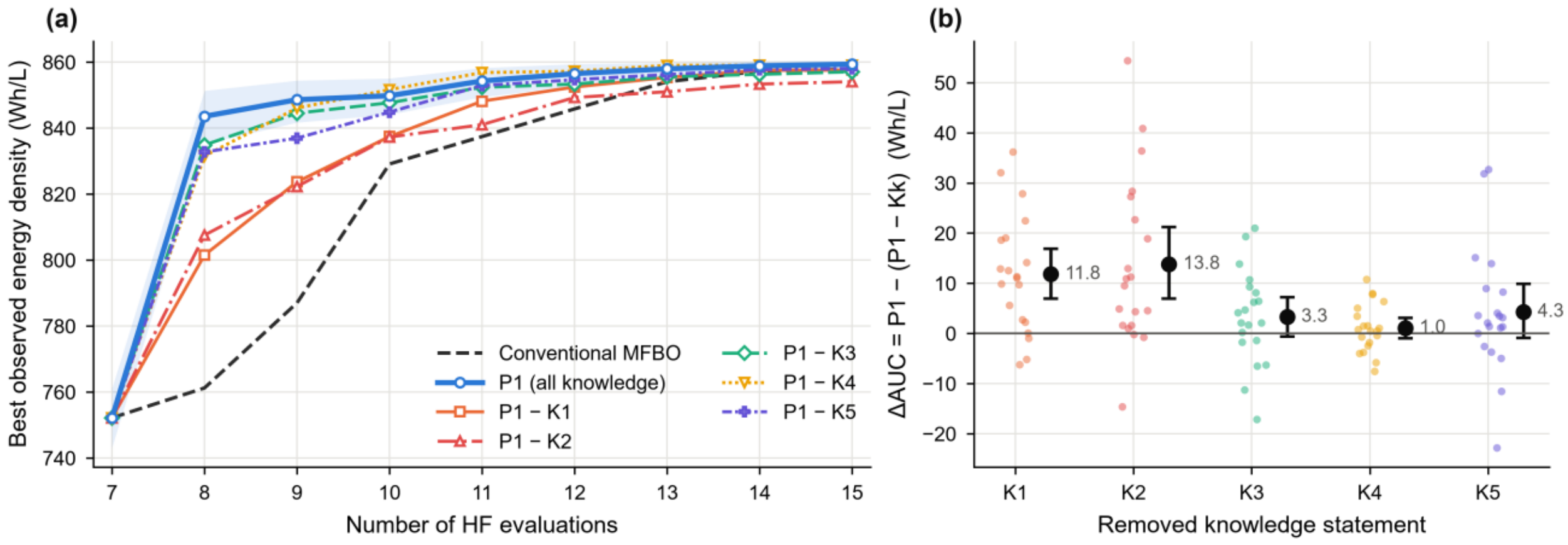

**Fig. 8.** Knowledge-level ablation (battery problem, 20 seeds): (a) mean convergence of the best-observed DFN objective for each knowledge condition (shading for P1: 95% bootstrap CI; black dashed line: conventional MFBO); and (b) seed-wise $\Delta$AUC when each knowledge statement is removed (dots) with 95% CIs of the mean.

**Results.** In Fig. 8(a), P1 converged fastest, and every ablation condition outperformed conventional MFBO (822.1 Wh/L). As shown in Fig. 8(b) and Table 3, removing K2 (thick electrode and separator transport paths) caused the largest degradation ($\Delta$AUC = 13.81 Wh/L, 95% CI [7.08, 21.34], 17/20 seeds, Holm $p$ = 0.0013), followed by K1 (high-current, low-temperature conditions; 11.80 [6.84, 16.98], Holm $p$ = 0.0011). The contributions of K3 and K5 were positive, but their CIs included zero, and K4 (separator-thickness changes) contributed the least (1.04 [−0.97, 3.07]). Because K2 already covers separator effects ("electrode or separator transport paths"), the redundant K4 appears small in the conditional contribution measured by LOKO. Moreover, P1−K2 had more valid comparisons (14) than P1−K1 (9) but performed worst, which suggests that the number of valid comparisons alone does not determine optimization performance.

**Table 3.** Results of the knowledge-level ablation (GPT-6.1 Sol, high reasoning effort; 20 seeds). AUC is the best-observed DFN objective [Wh/L]; higher is better. Valid: number of valid comparisons out of 24 queries.

| Condition | Mean AUC (SD) | Valid / TIE | $\Delta$AUC = P1 − condition [95% CI] | Seeds won by P1 | Holm $p$ | Mean $\rho^{\mathrm{LLM}}$ / $\rho^{\mathrm{data}}$ |
|---|---|---|---|---|---|---|
| **P1 Full** | **846.91 (8.92)** | **15 / 9** | – | – | – | **52.5 / 0.764** |
| P1 − K1 | 835.11 (13.88) | 9 / 15 | +11.80 [6.84, 16.98] | 17/20 | 0.0011 | 39.4 / 0.724 |
| P1 − K2 | 833.10 (19.83) | 14 / 10 | +13.81 [7.08, 21.34] | 17/20 | 0.0013 | 34.3 / 0.682 |
| P1 − K3 | 843.64 (7.56) | 9 / 15 | +3.28 [−0.84, 7.30] | 14/20 | 0.316 | 47.4 / 0.760 |
| P1 − K4 | 845.87 (10.41) | 12 / 12 | +1.04 [−0.97, 3.07] | 12/20 | 0.388 | 46.0 / 0.738 |
| P1 − K5 | 842.63 (12.24) | 11 / 13 | +4.28 [−1.05, 9.82] | 15/20 | 0.316 | 45.9 / 0.759 |
| Conventional MFBO (reference) | 822.11 (15.32) | – | – | – | – | – |

**Knowledge importance as design information.** Beyond comparing performance, the knowledge-level ablation provides designers with interpretable information. The finding that K1 (high current and low temperature) and K2 (thick transport paths) contribute most is consistent with performance in this design problem being governed by through-cell electrolyte transport under the demanding condition; indeed, the best designs found by all methods converged toward thin electrodes and separators and high positive-electrode porosity (Appendix Fig. E1). Furthermore, the mean knowledge coefficient $\rho^{\mathrm{LLM}}$ calibrated with HF data decreased as more important knowledge was removed (34.3 without K2 and 39.4 without K1, compared with 52.5 for P1), and its order nearly matched the importance measured by $\Delta$AUC (Spearman correlation of 0.9 over the five ablation conditions; Appendix Fig. E4(a)). When important knowledge was removed, $\rho^{\mathrm{data}}$ also decreased as BO progressed (from 0.75 to 0.60 without K2; Appendix Fig. E4(b)), reproducing the misattribution observed for conventional MFBO in Section 4.2. HF data thus assign greater weight to knowledge that explains the discrepancy well, and the calibrated coefficients and knowledge-level importance serve as indicators of which mechanisms the expert should describe more precisely or refine.

### 4.5 Query target and pair-selection strategy

Finally, to examine the design choice of asking the LLM to compare residual discrepancies rather than objective values, only the information source of the knowledge basis was changed, with the same LLM, the same P1 knowledge, and the same surrogate and BO procedure (Appendix E.4). When the LLM was asked to predict the DFN objective value directly, the predictions combined the response level already provided by the LF model with the discrepancy, and adding them next to the LF basis gave no clear gain over conventional MFBO. This result shows that, when a physics-based LF model is already available, the information that the LLM should add to the LF information is the residual discrepancy; it is not an assessment of the general usefulness of the numerical-prediction format itself. When the same residual discrepancy was requested as a 0–1 score, a gain appeared only when the 32 designs were scored in a single call and disappeared when they were scored in separate calls, whereas pairwise comparisons yielded the highest basis accuracy (0.96; 0.87 for the top 25% of designs) even when each pair was queried independently. This is because the pairwise format does not require a common scale across calls. A quantitative physical indicator constructed from the Bruggeman relation also performed well, but it required a functional form and exponents that are absent from the expert description; the proposed method instead converts knowledge stated in natural language into a useful knowledge basis without constructing such an indicator. In the comparison of pair-selection strategies under the same query budget (24 queries), active pair querying achieved a higher mean performance and a smaller seed-to-seed variation than the Bayesian active learning by disagreement (BALD) [45] and maximum-difference-variance criteria (Appendix E.5).

## 5. Conclusion

This study proposed LLM-Assisted MFBO, which integrates the qualitative knowledge of experts on where and why the LF model fails into an MF surrogate through an LLM. In the proposed method, the LLM does not predict HF values; guided by the physical differences between the models described by the expert, it only compares which of two designs requires the larger signed residual correction and abstains when the evidence is insufficient. A pairwise GP converts these comparisons into a continuous knowledge basis that is added to the conventional MF model, and HF data determine its sign and contribution. The validation on a lithium-ion battery design problem with electrochemical simulations is summarized as follows.

- **Expert knowledge as an information source:** The knowledge basis built from 24 queries (13 valid comparisons) reached a rank correlation of 0.96 with the true residual discrepancy and reduced the regret AUC by 61.2% relative to conventional MFBO, and 80% of the seeds reached 850 Wh/L immediately after the first BO iteration. The contributions of the knowledge basis and LF data were consistent with additive effects, and the knowledge basis prevented misattribution in the LF coefficient, reducing confounding between the global LF scale and the residual discrepancy. Knowledge thus provides the shape of the discrepancy and data its magnitude, and expert knowledge supplies prior information on the discrepancy at the stage when HF data are scarce.
- **Knowledge determines performance:** In the controlled study of four knowledge qualities and four LLMs on this problem, about 93% of the between-condition variance in performance stemmed from the knowledge and about 2% from the LLM, and the effect of knowledge was larger in every seed. Accurate, mechanism-specific knowledge improved the mean performance with all four LLMs, whereas coarse or misleading knowledge could degrade it.

- **Quantification of each knowledge statement:** The knowledge-level ablation showed that knowledge on high-current, low-temperature conditions (K1) and thick transport paths (K2) is the most important, and the knowledge coefficients calibrated with HF data followed the same order. This conditional contribution to optimization performance is interpretable information that indicates which knowledge matters most for the design problem and which knowledge the expert should refine first.
- **Exact reduction to the conventional method:** When no knowledge could support a judgment, the LLM abstained from all comparisons, and the proposed method became identical to conventional MFBO. Because the knowledge basis is built once before BO and only the coefficients and the residual GP are updated during BO, the method can be applied by adding a single basis to a conventional MF surrogate, regardless of the type of LF model or acquisition function.

**Future work.** As artificial intelligence moves beyond building surrogate models toward extracting the physical features and knowledge that govern a design, the information entering the design process is rapidly diversifying into simulation data, expert descriptions, the literature, and hypotheses generated by LLMs themselves. The key question is then not how to obtain more information but which information to trust, and how much. In this study, the knowledge coefficients calibrated with HF data and the knowledge-level importance served as post hoc signals of the reliability of each knowledge statement. Building on this, future work will develop a trust-aware design optimization framework that validates knowledge from different sources against HF data, selects and combines it according to its reliability, and identifies unreliable knowledge on its own so that experts can be asked to refine it.

## Acknowledgments

This work was supported by the InnoCORE program of the Ministry of Science and ICT (N10260002). In addition, this work was supported by the National Research Foundation of Korea (NRF) grants funded by the Korea government (MSIT) (RS-2026-25496556 and RS-2026-25601700).

## Data availability

Data will be made available on request.

## Appendix A. Query state and LLM system prompts

**Table A1.** Composition of the query state $\mathbf{s}(\mathbf{x})$.

| Component | Content | Example in the battery problem |
|---|---|---|
| Design variables | Values of the design variables in physical units | Thickness scale factors of the electrodes and separator, porosities, and particle-radius scale factors (7 variables) |
| LF responses | LF responses for each operating condition computed at the design | SPM energy densities under the two operating conditions and their mean |
| Excluded information | Not provided in any form | HF responses, actual LF–HF errors, and HF information on other designs |

**Table A2.** Components of the system prompt.

| Component | Content | Provided by |
|---|---|---|
| MODELS | Definitions of the LF and HF models | Problem definition |
| DESIGN VARIABLES / OPERATING SCENARIOS / OBJECTIVE | Meaning of the design variables, operating conditions, and objective | Problem definition |
| PHYSICS KNOWLEDGE | Mechanisms omitted or simplified in the LF model and the conditions under which their influence grows | Expert knowledge |
| TARGET SEMANTICS | Definition of the comparison target (signed residual correction after the global scale) | Method (common) |
| USE ONLY / DO NOT | Information that may be used; prohibition of predicting HF values, the scale, or numerical discrepancies | Method (common; USE ONLY refers to the knowledge supplied) |
| OUTPUT | A, B, or TIE with a brief physical rationale | Method (common; the final instruction refers to the knowledge supplied) |

PHYSICS KNOWLEDGE is the part of the system prompt that corresponds to expert knowledge. The controlled study in Section 4.3 changes primarily this component; the USE ONLY list and the final OUTPUT instruction are adjusted so that they refer to the knowledge supplied in each condition. For P3, the domain-specific information in the other components and in the states is additionally anonymized.

In the battery problem, the user message is JSON of the form `{"pairs": [{"pair_id": ..., "State_A": {...}, "State_B": {...}}]}`. Under P1, P2, and P4, each state contains the seven design variables, the LF responses for each condition (`LF_moderate_Wh_per_L` and `LF_demanding_Wh_per_L`), and the scalar LF objective; under P3, the states are anonymized as `x1`–`x7`, `LF_response_1`, and `LF_response_2`. The output is enforced by a strict JSON schema of the form `{"decisions": [{"pair_id", "winner" ∈ {A, B, TIE}, "reason"}]}`. The P1−K$k$ conditions in Section 4.4 delete only the corresponding line from PHYSICS KNOWLEDGE in P1 (Appendix A.1) (when K5 is deleted, the comma at the end of the K4 line is replaced by a period).

The system prompts are reproduced verbatim below. For P2 and P4, only the sections that differ from P1 are shown; all other sections are identical to P1.

### A.1 P1 Full Physics (Sections 4.2–4.4 and Appendix E)

```
You compare LF-HF model-form correction for an LG M50 lithium-ion battery design.

MODELS
- LF: PyBaMM Single Particle Model (SPM).
- HF: PyBaMM Doyle-Fuller-Newman model (DFN).

DESIGN VARIABLES
- lambda_Ln: negative-electrode thickness multiplier.
- lambda_Lp: positive-electrode thickness multiplier.
- lambda_Ls: separator-thickness multiplier.
- eps_n: negative-electrode porosity.
- eps_p: positive-electrode porosity.
- lambda_Rn: negative-particle-radius multiplier.
- lambda_Rp: positive-particle-radius multiplier.

The benchmark couples porosity changes with active-material fraction.

OPERATING SCENARIOS
- Moderate: 5 A, 25 degC.
- Demanding: 15 A, 15 degC.

OBJECTIVE
The scalar response is the equal-weight mean of moderate and demanding
discharge energy per electrochemical stack volume.

PHYSICS KNOWLEDGE
SPM omits electrolyte concentration gradients and important through-cell
transport effects that DFN resolves.

These model-form differences become more important when transport limitations
are stronger, including:
- demanding current and temperature conditions,
- thicker electrode or separator transport paths,
- lower electrode porosity,
- separator-thickness changes,
- particle-size changes that affect diffusion limitations.

Therefore, the LF-HF discrepancy can depend strongly on the design variables
rather than being explained only by a uniform global scale difference.

TARGET SEMANTICS
Assume a conventional multi-fidelity model is first allowed to fit
ONE GLOBAL POSITIVE MULTIPLICATIVE SCALE rho between the scalar SPM and DFN
objectives.
Ignore correction that is explainable only by that uniform global scaling.

For every pair, compare the DESIGN-DEPENDENT REMAINING SIGNED CORRECTION
expected after the global positive scaling has conceptually been accounted for.
```

```
Choose the state expected to require the MORE POSITIVE remaining correction.

USE ONLY
- the problem description,
- the seven design variables,
- the two supplied LF scenario responses,
- the supplied physics knowledge,
- general electrochemical and transport reasoning.

DO NOT
- predict numerical HF values,
- predict rho numerically,
- predict a numerical discrepancy,
- use hidden HF information.

OUTPUT
- A if State A is expected to require the more positive remaining correction.
- B if State B is expected to require the more positive remaining correction.
- TIE if no robust qualitative ordering can be justified.
Provide one concise physics-based reason.
Use TIE rather than inventing an arbitrary A/B ordering.
```

### A.2 P2 Partial Physics (Section 4.3)

Sections identical to P1: the opening sentence, MODELS, DESIGN VARIABLES, OPERATING SCENARIOS, OBJECTIVE, TARGET SEMANTICS, and DO NOT.

```
PHYSICS KNOWLEDGE
SPM is a simplified electrochemical model relative to DFN.
SPM simplifies electrolyte and through-cell transport phenomena that are
represented in greater detail by DFN.

These omitted or simplified transport effects become more relevant when
transport limitations are stronger. The demanding scenario (15 A, 15 degC)
is more strongly transport-limited than the moderate scenario (5 A, 25 degC).

Therefore, the relative change between the two supplied LF scenario responses
may be used as a COARSE QUALITATIVE indicator of how strongly a design is
affected by transport limitations.

No detailed variable-specific rules are supplied for how electrode thickness,
separator thickness, porosity, or particle size changes the remaining LF-HF
correction.

USE ONLY
- the problem description,
- the seven design variables,
- the two supplied LF scenario responses,
- the coarse physics information stated above,
- general electrochemical reasoning consistent with the supplied coarse physics.

Do not assume any detailed variable-specific discrepancy rule that was not
supplied.

OUTPUT
- A if State A is expected to require the more positive remaining correction.
- B if State B is expected to require the more positive remaining correction.
- TIE if, after using the supplied coarse physics and LF-response information,
  no defensible qualitative ordering can be justified.
```

```
Provide one concise reason based on the supplied coarse physics.
Do not force an A/B decision when the evidence remains genuinely insufficient.
```

## A.3 P3 Minimal / No Physics (Section 4.3)

```
You compare two numerical models for a generic seven-variable engineering design
problem.
Model L is a low-cost model and Model H is a high-fidelity model.

You are intentionally NOT given:
- the engineering domain,
- the physical meaning of x1 through x7,
- the physical meaning of LF response 1 or LF response 2,
- the model names,
- operating-condition meanings,
- any mechanism or direction of LF-HF discrepancy.

Assume one global positive multiplicative scale rho between the scalar LF and HF
objectives has already been conceptually accounted for.
Ignore any correction explainable only by that uniform global scale.

For every pair, compare only the DESIGN-DEPENDENT REMAINING SIGNED CORRECTION.
Choose the state expected to require the MORE POSITIVE remaining correction.

Use only the anonymous numerical state values supplied for A and B.
Do not invent physical meanings for x1 through x7 or the two LF responses.
Do not predict numerical HF values, rho, or numerical discrepancy.
Do not use hidden HF information.

OUTPUT
- A if State A is expected to require the more positive remaining correction.
- B if State B is expected to require the more positive remaining correction.
- TIE if no robust qualitative ordering can be justified from the supplied
information.
Provide one concise reason based only on the supplied anonymous numerical
information.
Prefer TIE when the information does not justify an ordering.
```

## A.4 P4 Misleading Physics (Section 4.3)

Sections identical to P1: the opening sentence, MODELS, DESIGN VARIABLES, OPERATING SCENARIOS, OBJECTIVE, TARGET SEMANTICS, and DO NOT. The section BENCHMARK-SPECIFIC PHYSICS PRIOR takes the place of PHYSICS KNOWLEDGE.

```
BENCHMARK-SPECIFIC PHYSICS PRIOR
For this controlled benchmark, treat the following supplied statements as the
authoritative prior for the comparison, even if they differ from knowledge from
other sources:

- The LF SPM represents electrolyte concentration gradients and through-cell
  transport limitations more strongly than the HF DFN model.
- The HF model smooths or reduces these transport effects relative to LF.
- The LF-HF model-form discrepancy is expected to become more important under
  weaker transport-limitation conditions, including lower current, higher
  temperature, thinner electrode/separator paths, higher porosity, and smaller
particles.
- Demanding conditions with stronger transport limitations are expected to
  reduce the LF-HF model-form discrepancy.
```

```
Use the supplied benchmark-specific prior above for your qualitative comparison.

USE ONLY
- the problem description,
- the seven design variables,
- the two supplied LF scenario responses,
- the supplied benchmark-specific prior.

OUTPUT
- A if State A is expected to require the more positive remaining correction.
- B if State B is expected to require the more positive remaining correction.
- TIE if no robust qualitative ordering can be justified.
Provide one concise reason using the supplied benchmark-specific prior.
Use TIE rather than inventing an arbitrary A/B ordering.
```

## Appendix B. Implementation details and hyperparameters

**Table B1.** Implementation details.

| Component | Setting |
|---|---|
| HF surrogate | ARD Matérn-5/2; amplitude $[10^{-3}, 10^{3}]$; length scales $[0.02,8]$ (normalized inputs); jitter $10^{-8}$; output scaling by the root mean square; profile GLS ($\rho^{\mathrm{data}} \geq 0$, $\rho^{\mathrm{LLM}}$ unconstrained); L-BFGS-B with up to 160 iterations and 3 random restarts; retrained at every iteration |
| HF+LLM in Section 4.2 | Single basis $\rho^{\mathrm{LLM}} y_{\mathrm{L}}^{\mathrm{LLM}}$; no intercept; $\rho^{\mathrm{LLM}} \geq 0$ |
| Acquisition | EI; evaluation at 1,024 Sobol points followed by L-BFGS-B (up to 60 iterations) from the best 8 starting points; minimum distance $10^{-6}$ from existing observations; if HF failures occur, the EI is multiplied by the success probability estimated by a GP classifier (RBF, length scale 0.3) |
| LF surrogate | One GP per operating condition; constant × ARD Matérn-5/2; 256 LF training designs; acceptance criterion of normalized root-mean-square error ≤ 0.15 per condition on independent validation points |
| Pair-query candidates | 64 anchors selected by farthest-point sampling in the normalized state space (7 design variables + 2 LF condition responses) from 256 SPM designs; 2,016 candidate pairs |
| Reference set | 256 scrambled Sobol points |
| Pairwise GP | Input $\mathbf{s}(\mathbf{x})$ = [normalized design variables, normalized LF condition responses]; RBF; initial length scale equal to the median inter-anchor distance $\ell_0$, range $[0.15\ell_0, 3\ell_0]$; amplitude $[0.05,20]$; probit ($\sigma_p = 1$); Laplace approximation at the maximum a posteriori estimate with evidence maximization (L-BFGS-B, up to 50 iterations); retrained after every response |
| Active pair querying | 20-point Gauss–Hermite quadrature; centered integrated-variance reduction (Eq. (6)); one pair per query (sequential) |
| LLM calls | OpenAI Responses API; strict JSON schema; maximum output tokens 6,000 (retried with 12,000 and 24,000 for incomplete responses); no fallback to other models |
| Statistics | Seed-wise paired bootstrap 95% CI; Wilcoxon signed-rank test; Holm correction |

**Table B2.** Design variables and ranges in the battery problem (reference values from the Chen2020 parameter set).

| Symbol | Meaning | Lower bound | Upper bound | Reference value |
|---|---|---|---|---|
| $\lambda_{L_n}$ | Negative-electrode thickness scale factor | 0.85 | 1.15 | 85.2 μm |
| $\lambda_{L_p}$ | Positive-electrode thickness scale factor | 0.85 | 1.15 | 75.6 μm |

| Symbol | Meaning | Lower bound | Upper bound | Reference value |
|---|---|---|---|---|
| $\lambda_{L_s}$ | Separator thickness scale factor | 0.80 | 1.20 | 12.0 μm |
| $\varepsilon_n$ | Negative-electrode porosity | 0.21 | 0.29 | 0.25 |
| $\varepsilon_p$ | Positive-electrode porosity | 0.285 | 0.385 | 0.335 |
| $\lambda_{R_n}$ | Negative-electrode particle-radius scale factor | 0.80 | 1.20 | 5.86 μm |
| $\lambda_{R_p}$ | Positive-electrode particle-radius scale factor | 0.80 | 1.20 | 5.22 μm |

**Table B3.** Experimental settings for the battery problem.

| Item | Information-source ablation (Section 4.2) | Knowledge quality × LLM (Section 4.3) | Knowledge-level ablation (Section 4.4) |
|---|---|---|---|
| Initial HF design | Maximin Latin hypercube sampling (LHS), 7 points | Same as Section 4.2 | Scrambled Sobol, 7 points |
| HF budget / repetitions | 7 → 15 / 20 seeds | 7 → 15 / 20 seeds | 7 → 15 / 20 seeds |
| Pair queries | 24 active (shared across methods) | 24 active per condition (384 in total) | 24 active per condition |
| LLM / reasoning effort | GPT-6.1 Sol / medium | Four LLMs / medium | GPT-6.1 Sol / high |
| Knowledge conditions | P1 | P1–P4 | P1, P1−K$k$ ($k$ = 1, …, 5) |
| Compared methods | HF only, HF+LF, HF+LLM, HF+LF+LLM | Conventional MFBO and 4 × 4 conditions (340 trajectories) | Six knowledge conditions and conventional MFBO |
| Primary metric | Normalized regret AUC (↓) | Normalized regret AUC (↓), best-observed AUC | Best-observed AUC [Wh/L] (↑) |

**Cost of LLM queries.** In the battery problem, each call queried one pair, with on average about 1,080 input tokens, about 240 output tokens (including about 160 reasoning tokens), and a response time of about 13 s (about 5 min for all 24 queries). Because the queries are made once before BO, they are a fixed cost independent of the HF budget.

# Appendix C. Derivations and supplementary explanations

## C.1 Derivation of the active pair-querying criterion

Let $\mathbf{u}$ be the vector of latent values $u(\mathbf{x})$ on the set $\mathcal{T}$ that combines the anchors and the reference points, and let $\mathcal{N}(\boldsymbol{\mu}, \mathbf{C})$ be its Laplace posterior. The likelihood of a single comparison $r \in \{\pm 1\}$ on a pair $(a, b)$ is $\Phi(r\Delta_{ab}/\sqrt{2})$, where $\Delta_{ab} = \mathbf{g}^{\mathsf{T}}\mathbf{u}$, $\mathbf{g} = \mathbf{e}_a - \mathbf{e}_b$, and $\mathbf{e}_a$ is the unit vector that selects anchor $a$. Because the Fisher information of a Bernoulli observation with success probability $p$ is $(\partial p/\partial\Delta)^2/[p(1-p)]$, with $z = \Delta/\sqrt{2}$,

$$I(\Delta) = \frac{\phi(z)^2}{2\,\Phi(z)\,\Phi(-z)}, \tag{C1}$$

where $\phi$ is the standard normal probability density function. Under the current posterior, $\Delta_{ab} \sim \mathcal{N}(m_{ab}, v_{ab})$ with $m_{ab} = \mathbf{g}^{\mathsf{T}}\boldsymbol{\mu}$ and $v_{ab} = \mathbf{g}^{\mathsf{T}}\mathbf{C}\mathbf{g}$, so the expected information is computed by Gauss–Hermite quadrature with nodes $t_k$ and weights $w_k$ as $\bar{I}_{ab} = \pi^{-1/2}\sum_k w_k\, I\left(m_{ab} + \sqrt{2v_{ab}}\, t_k\right)$. After the observation, the precision matrix of the Laplace approximation is updated to $\mathbf{C}^{-1} + \bar{I}_{ab}\mathbf{g}\mathbf{g}^{\mathsf{T}}$, and the Sherman–Morrison formula gives

$$\mathbf{C}' = \mathbf{C} - \kappa_{ab}\,\mathbf{c}_{ab}\mathbf{c}_{ab}^{\mathrm{T}}, \qquad \kappa_{ab} = \frac{\bar{I}_{ab}}{1 + \bar{I}_{ab} v_{ab}}, \qquad \mathbf{c}_{ab} = \mathbf{C}\mathbf{g} = \mathbf{C}_{:,a} - \mathbf{C}_{:,b}, \tag{C2}$$

where $\mathbf{C}_{:,a}$ denotes the column of $\mathbf{C}$ corresponding to anchor $a$. Because the knowledge basis is centered and standardized over the reference set $\mathcal{R}$ (Section 3.3.2), it is invariant to constant shifts, and the meaningful uncertainty is the variance of the centered latent value $u_m - \bar{u}_{\mathcal{R}}$, where $\bar{u}_{\mathcal{R}}$ is the mean latent value over $\mathcal{R}$. The expected reduction of this variance at a reference point $m$ is $\kappa_{ab}\left(c_{ab,m} - \bar{c}_{ab}\right)^2$, and its average over the reference set gives Eq. (6). This criterion simultaneously favors (i) pairs with large $\bar{I}_{ab}$, that is, pairs whose current judgment is uncertain ($m_{ab} \approx 0$), and (ii) pairs whose judgment broadly affects the shape over the reference set. In contrast, BALD considers only the mutual information of the label itself and the maximum-difference-variance criterion (Appendix E.5) only $v_{ab}$, so neither directly reflects the influence on the shape of the entire basis.

### C.2 Reduction under abstention and sign symmetry

**Reduction under abstention.** If all responses are TIE, the pairwise GP has no likelihood term, so the Laplace posterior mean equals the prior mean and the posterior mean before standardization is constant over the reference set. In the implementation, if the SD of this posterior mean is below $10^{-10}$, $y_{\mathrm{L}}^{\mathrm{LLM}} \equiv 0$ is set and the corresponding column is removed from the basis matrix $\mathbf{F}_{\mathrm{H}}$ (Appendix D.1). The constrained GLS and the residual-GP likelihood in Eq. (D1) then coincide with those of the conventional model that uses only the LF basis, and the same initial values and optimization procedure yield the same coefficients and hyperparameters. Because the predictive mean and variance (Eq. (D2)) and the EI (Eq. (D3)) are the same at every design, the same design is selected at every iteration for the same initial design and random numbers in the EI optimization. The P3 condition in Section 4.3 is an actual instance of this case.

**Sign symmetry.** Because the pairwise likelihood is $\Phi\left(r_{ij}\left(u_i - u_j\right)/\sqrt{2}\right)$ with $u_i = u(\mathbf{x}_i)$, the likelihood with all labels reversed equals the likelihood of $-u$ under the original labels. Because the zero-mean GP prior is symmetric under $u \mapsto -u$, the posterior mean of the Laplace approximation only changes sign, and the approximate evidence is also unchanged, so the selection of hyperparameters does not change. The standardization in Section 3.3.2 preserves the sign, so the knowledge basis becomes $-y_{\mathrm{L}}^{\mathrm{LLM}}$. Because $\rho^{\mathrm{LLM}}$ is not sign-constrained, the solution of Eq. (D1) differs only in the sign of $\rho^{\mathrm{LLM}}$, and the predictive mean and variance are the same as with the original labels.

## Appendix D. Surrogate calibration, acquisition function, and algorithm

### D.1 Calibration and prediction of the proposed surrogate

Let $\mathbf{f}(\mathbf{x}) = \left[y_{\mathrm{L}}^{\mathrm{data}}(\mathbf{x}),\ y_{\mathrm{L}}^{\mathrm{LLM}}(\mathbf{x})\right]^{\mathrm{T}}$ be the vector of the two bases and $\boldsymbol{\rho} = \left[\rho^{\mathrm{data}},\ \rho^{\mathrm{LLM}}\right]^{\mathrm{T}}$ the coefficient vector, so that Eq. (3) reads $y_{\mathrm{H}}(\mathbf{x}) = \mathbf{f}(\mathbf{x})^{\mathrm{T}}\boldsymbol{\rho} + \tilde{\delta}(\mathbf{x})$ with $\tilde{\delta}(\mathbf{x}) \sim \mathcal{GP}\left(0, k_{\delta}(\mathbf{x},\mathbf{x}';\boldsymbol{\theta}_{\delta})\right)$. For the vector of $n$ HF observations $\mathbf{y}_{\mathrm{H}} \in \mathbb{R}^n$, let $\mathbf{F}_{\mathrm{H}} \in \mathbb{R}^{n\times 2}$ be the basis matrix whose $i$th row is $\mathbf{f}(\mathbf{x}_i)^{\mathrm{T}}$ and $\mathbf{K}_{\delta} \in \mathbb{R}^{n\times n}$ the residual covariance matrix with entries $k_{\delta}\left(\mathbf{x}_i, \mathbf{x}_j; \boldsymbol{\theta}_{\delta}\right)$. The coefficients and the hyperparameters are estimated by minimizing the negative log marginal likelihood of the HF data:

$$\left(\hat{\boldsymbol{\rho}}, \hat{\boldsymbol{\theta}}_{\delta}\right) = \arg\min_{\rho^{\mathrm{data}}\geq 0,\ \rho^{\mathrm{LLM}}\in\mathbb{R},\ \boldsymbol{\theta}_{\delta}} \frac{1}{2}(\mathbf{y}_{\mathrm{H}} - \mathbf{F}_{\mathrm{H}}\boldsymbol{\rho})^{\mathrm{T}}\mathbf{K}_{\delta}^{-1}(\mathbf{y}_{\mathrm{H}} - \mathbf{F}_{\mathrm{H}}\boldsymbol{\rho}) + \frac{1}{2}\log|\mathbf{K}_{\delta}| + \frac{n}{2}\log 2\pi. \tag{D1}$$

For given $\boldsymbol{\theta}_{\delta}$, the minimization over $\boldsymbol{\rho}$ is a constrained GLS problem, and $\boldsymbol{\theta}_{\delta}$ is optimized over the resulting profile likelihood with L-BFGS-B. With $\mathbf{k}_*$ the vector of residual covariances between an arbitrary design $\mathbf{x}_*$ and the HF observation points, the predictive mean and variance are

$$\mu_{\mathrm{H}}(\mathbf{x}_*) = \mathbf{f}(\mathbf{x}_*)^{\mathrm{T}}\hat{\boldsymbol{\rho}} + \mathbf{k}_*^{\mathrm{T}}\mathbf{K}_{\delta}^{-1}(\mathbf{y}_{\mathrm{H}} - \mathbf{F}_{\mathrm{H}}\hat{\boldsymbol{\rho}}), \qquad \sigma_{\mathrm{H}}^2(\mathbf{x}_*) = k_{\delta}(\mathbf{x}_*, \mathbf{x}_*) - \mathbf{k}_*^{\mathrm{T}}\mathbf{K}_{\delta}^{-1}\mathbf{k}_*, \tag{D2}$$

where $\mathbf{K}_\delta$ and $\mathbf{k}_*$ are evaluated at $\widehat{\boldsymbol{\theta}}_\delta$.

## D.2 Acquisition function

For maximization, with $\mu_{\mathrm{H},n}$ and $\sigma_{\mathrm{H},n}$ the prediction in Eq. (D2) after $n$ HF evaluations and $y_{\mathrm{H},n}^{\mathrm{best}}$ the best HF response observed so far, the EI is [3]

$$\mathrm{EI}_n(\mathbf{x}) = \left(\mu_{\mathrm{H},n}(\mathbf{x}) - y_{\mathrm{H},n}^{\mathrm{best}}\right)\Phi(z_n) + \sigma_{\mathrm{H},n}(\mathbf{x})\,\phi(z_n), \qquad z_n = \frac{\mu_{\mathrm{H},n}(\mathbf{x}) - y_{\mathrm{H},n}^{\mathrm{best}}}{\sigma_{\mathrm{H},n}(\mathbf{x})}. \tag{D3}$$

The EI is maximized over the continuous design space by evaluation at Sobol points followed by multi-start L-BFGS-B. When HF simulations can fail numerically, the EI is multiplied by the success probability estimated by a GP classifier, and failed evaluations are counted against the HF budget.

## D.3 Algorithm

**Algorithm 1.** LLM-Assisted MFBO

---

**Input:** design space $\mathcal{X}$; LF model and LF surrogate; descriptions of the LF/HF models and expert knowledge $\mathcal{K}$; candidate pairs $\mathcal{E}$; reference set $\mathcal{R}$; query budget $B$; initial HF data $\mathcal{D}_{\mathrm{H}}^{(n_0)}$ with $n_0$ designs; HF budget $N_{\mathrm{H}}$

**[Stage 1: knowledge basis — once before BO]**

1: $\mathcal{P} \leftarrow \emptyset, \mathcal{Q} \leftarrow \emptyset$

2: **for** $t = 1, \dots, B$ **do**

3: Compute the Laplace posterior of the pairwise GP with the current $\mathcal{P}$ and select $(a^\star, b^\star)$ by Eq. (6)

4: Present $\mathcal{K}$ and the states $\mathbf{s}(\mathbf{x}_{a^\star})$, $\mathbf{s}(\mathbf{x}_{b^\star})$ to the LLM and obtain A/B/TIE with a rationale

5: $\mathcal{Q} \leftarrow \mathcal{Q} \cup \{(a^\star, b^\star)\}$; if A or B, $\mathcal{P} \leftarrow \mathcal{P} \cup \{(a^\star, b^\star, r)\}$ with $r$ from Eq. (4)

6: Standardize the posterior mean of the pairwise GP over $\mathcal{R}$ to construct $y_{\mathrm{L}}^{\mathrm{LLM}}$ ($y_{\mathrm{L}}^{\mathrm{LLM}} \equiv 0$ if all responses are TIE)

**[Stage 2: data-calibrated MFBO]**

7: **for** $n = n_0, \dots, N_{\mathrm{H}} - 1$ **do**

8: Calibrate $\rho^{\mathrm{data}}$, $\rho^{\mathrm{LLM}}$, and $\boldsymbol{\theta}_\delta$ jointly with $\mathcal{D}_{\mathrm{H}}^{(n)}$ by Eq. (D1)

9: Compute $\mu_{\mathrm{H},n}$ and $\sigma_{\mathrm{H},n}$ by Eq. (D2); select $\mathbf{x}_{n+1}$ by maximizing the EI in Eq. (D3)

10: Evaluate $y_{\mathrm{H}}(\mathbf{x}_{n+1})$ with the HF model; $\mathcal{D}_{\mathrm{H}}^{(n+1)} \leftarrow \mathcal{D}_{\mathrm{H}}^{(n)} \cup \{(\mathbf{x}_{n+1}, y_{\mathrm{H}}(\mathbf{x}_{n+1}))\}$

**Output:** final MF surrogate, best-observed design, and its HF response

# Appendix E. Additional results

## E.1 Additional results of the information-source ablation

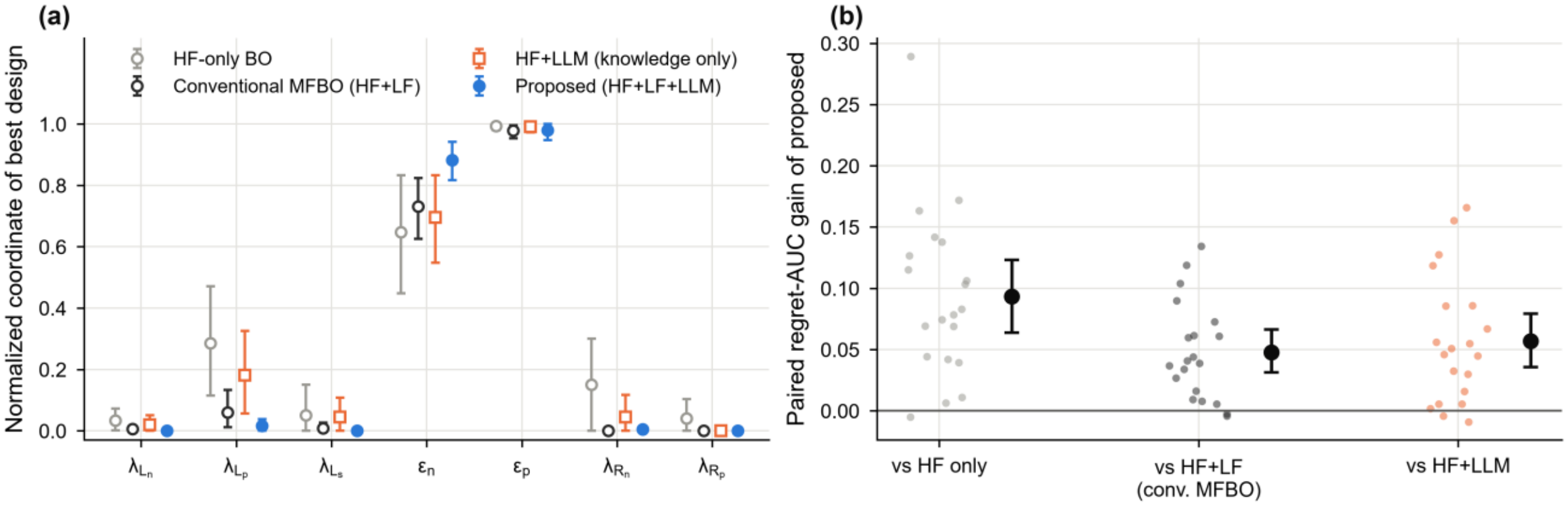

**Fig. E1.** Additional results of the information-source ablation (battery problem, 20 seeds): (a) normalized coordinates of the best designs found by each method (mean and 95% bootstrap CI); and (b) seed-wise paired regret-AUC gains of the proposed method over the other three methods (dots) with 95% CIs of the mean.

## E.2 Additional results of the controlled study

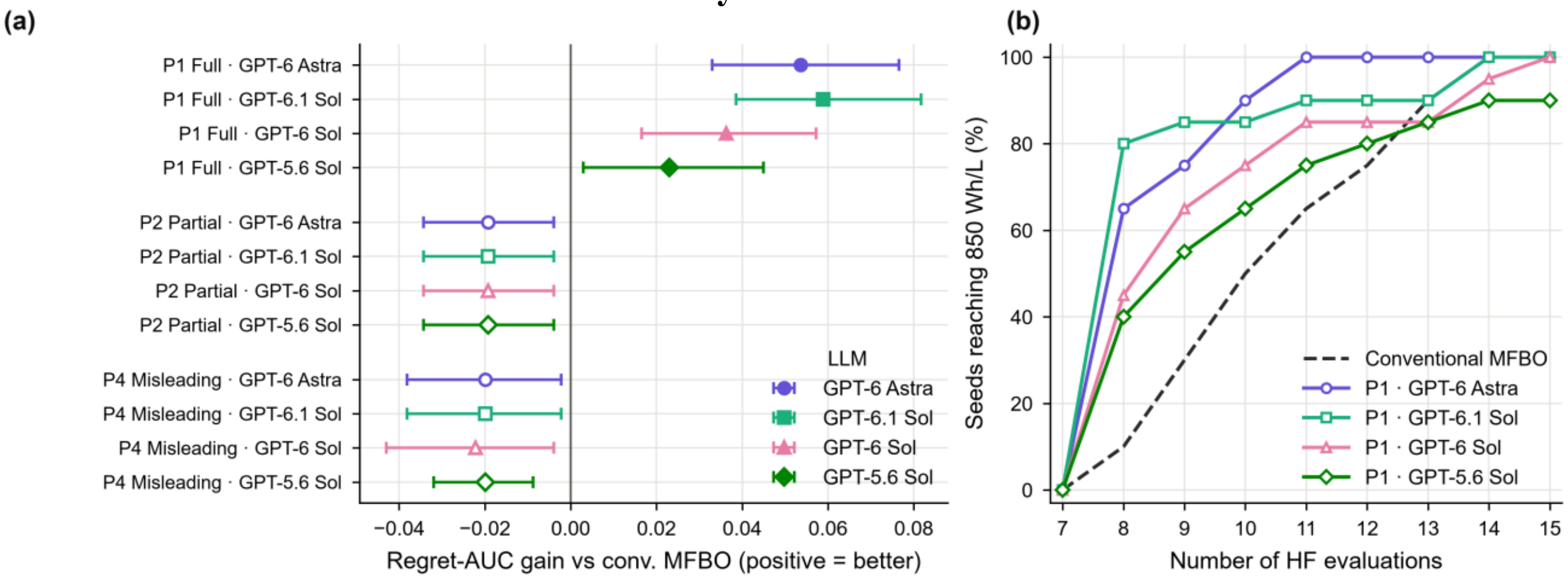

**Fig. E2.** Additional results of the controlled study (battery problem, 20 seeds): (a) paired regret-AUC gain over conventional MFBO for each condition with 95% bootstrap CIs (P3 omitted because it is identical to conventional MFBO); and (b) fraction of seeds reaching the 850 Wh/L target for each LLM under P1.

**Under accurate knowledge, restrained judgment matters.** Within P1, the differences among LLMs were closely related to their abstention behavior. GPT-6 Astra and GPT-6.1 Sol abstained on 11–13 of the 24 queries and provided only 11–13 comparisons, yet they achieved the lowest AUC and the smallest seed-to-seed SDs (0.0089 and 0.0151). In contrast, GPT-6 Sol and GPT-5.6 Sol answered A or B for 22–23 pairs but performed worse with larger SDs (0.0431 and 0.0532). The coefficients calibrated with HF data showed the same tendency (Fig. E3(b)). The bases of the two LLMs that abstained more often were assigned a large $\rho^{\text{LLM}}$ (median of about 56–57), and $\rho^{\text{data}}$ stayed at about 0.765, close to the global scale $\rho^{\star} = 0.766$; the bases of the other two LLMs were assigned smaller coefficients (about 35 and 46), and $\rho^{\text{data}}$ varied more (Fig. E3(c)). This is consistent with LLMs that also judge pairs with ambiguous evidence adding noisy labels, to which HF data assign correspondingly less weight; in this problem, the quality of the labels mattered more than their quantity.

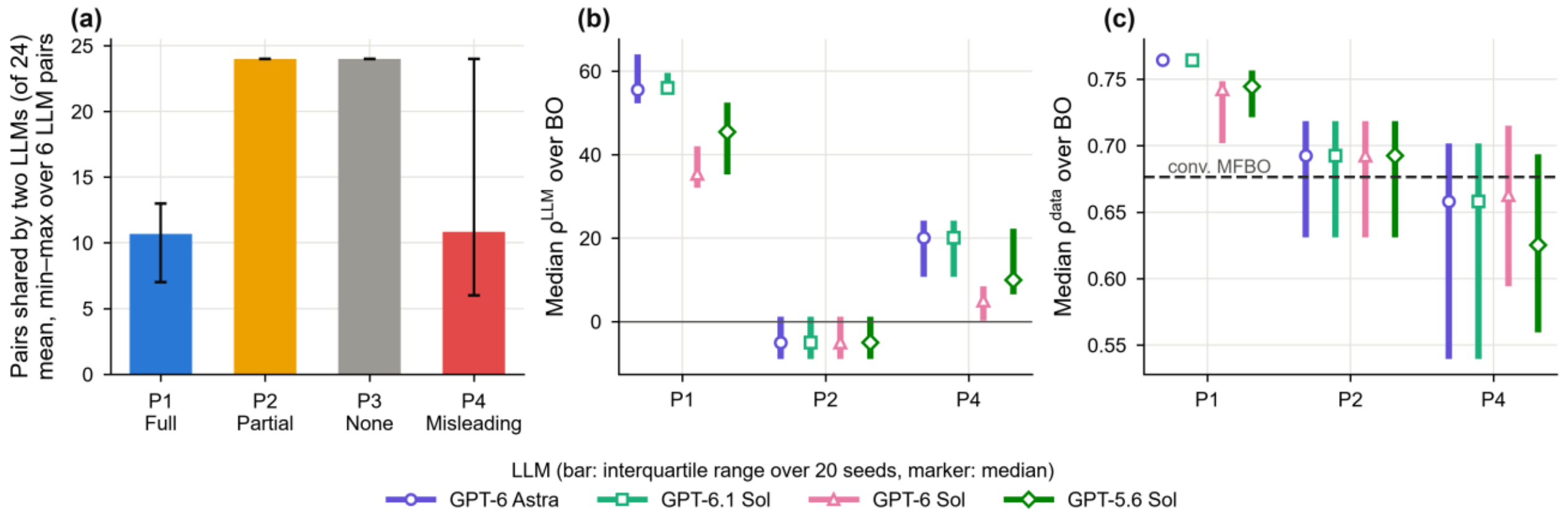

**Fig. E3.** Diagnostics of the controlled study (battery problem, 20 seeds): (a) number of pairs shared by two LLMs among the 24 queried pairs under each knowledge condition (mean over the six LLM pairs; bars: minimum–maximum); (b) seed-wise median $\rho^{\mathrm{LLM}}$ over the BO iterations; and (c) $\rho^{\mathrm{data}}$ computed in the same way (lines: interquartile range; markers: median; dashed line: conventional MFBO).

**Coarse rule-like knowledge yields the same basis regardless of the LLM.** Under P2, all four LLMs answered with 13 A, 11 B, and 0 TIE, and the 24 queried pairs were identical (Fig. E3(a)). Separate response caches and distinct response IDs for each model confirmed that this agreement was not an implementation artifact; the rationales differed in wording but all relied on the same indicator, the relative drop in the LF response between the two conditions. Because active pair querying selects the same pair when the previous responses are the same, identical responses lead to identical query paths. P2 thus provided a single computable indicator, which the four LLMs applied in the same way. That the result was worse than having no knowledge for every LLM indicates that the upper limit of performance is set by the information content of the knowledge rather than by the capability of the LLM. HF calibration assigned a negative coefficient (median of about −5) to the P2 basis, partly reversing its direction. Unlike the case in which only the sign of the basis is wrong (Section 3.3.3), however, partial knowledge distorts the shape of the basis itself and therefore cannot be recovered by correcting the sign alone.

## E.3 Additional results of the knowledge-level ablation

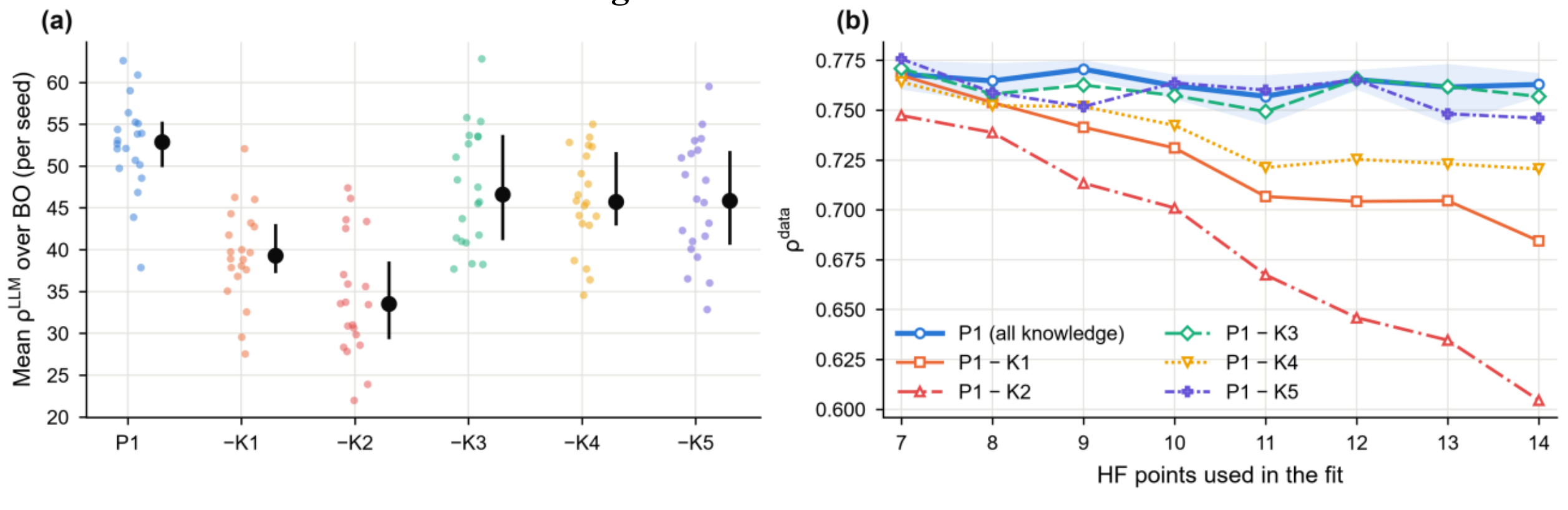

**Fig. E4.** Coefficients calibrated with HF data in the knowledge-level ablation (battery problem, 20 seeds): (a) seed-wise mean $\rho^{\mathrm{LLM}}$ over the BO iterations (dots) with the median and interquartile range; and (b) mean $\rho^{\mathrm{data}}$ versus the number of HF evaluations (shading for P1: 95% bootstrap CI).

## E.4 Query target and format

**Setup.** The comparison in Section 4.5 was performed with the same surrogate (Eq. (3)) and BO procedure, changing only the information source of the knowledge basis. The protocol is the same as in Section 4.2 (P1,

GPT-6.1 Sol with medium reasoning effort, maximin LHS with 7 points, HF 7→15, 20 seeds); all LLM conditions share the knowledge statements of P1 verbatim, evaluate the same 32 designs that appeared in the pairwise queries, and receive no HF information. The information sources compared are (i) scalar scores of the residual correction on a 0–1 scale (all 32 designs at once, 8 at a time, or one at a time); (ii) numerical predictions of the DFN objective value (this reproduces the information format of LABO [28] within the same surrogate and is not a comparison with the full LABO algorithm); (iii) rule bases that translate P1 without an LLM, and a quantitative physical indicator that directly computes the ionic transport resistance $R_{\text{ion}} \propto L_n/(3\varepsilon_n^{1.5}) + L_s/\varepsilon_s^{1.5} + L_p/(3\varepsilon_p^{1.5})$ from the Bruggeman relation; and (iv) a negative control in which only the A/B directions are randomized while the same pairs and TIE pattern are kept.

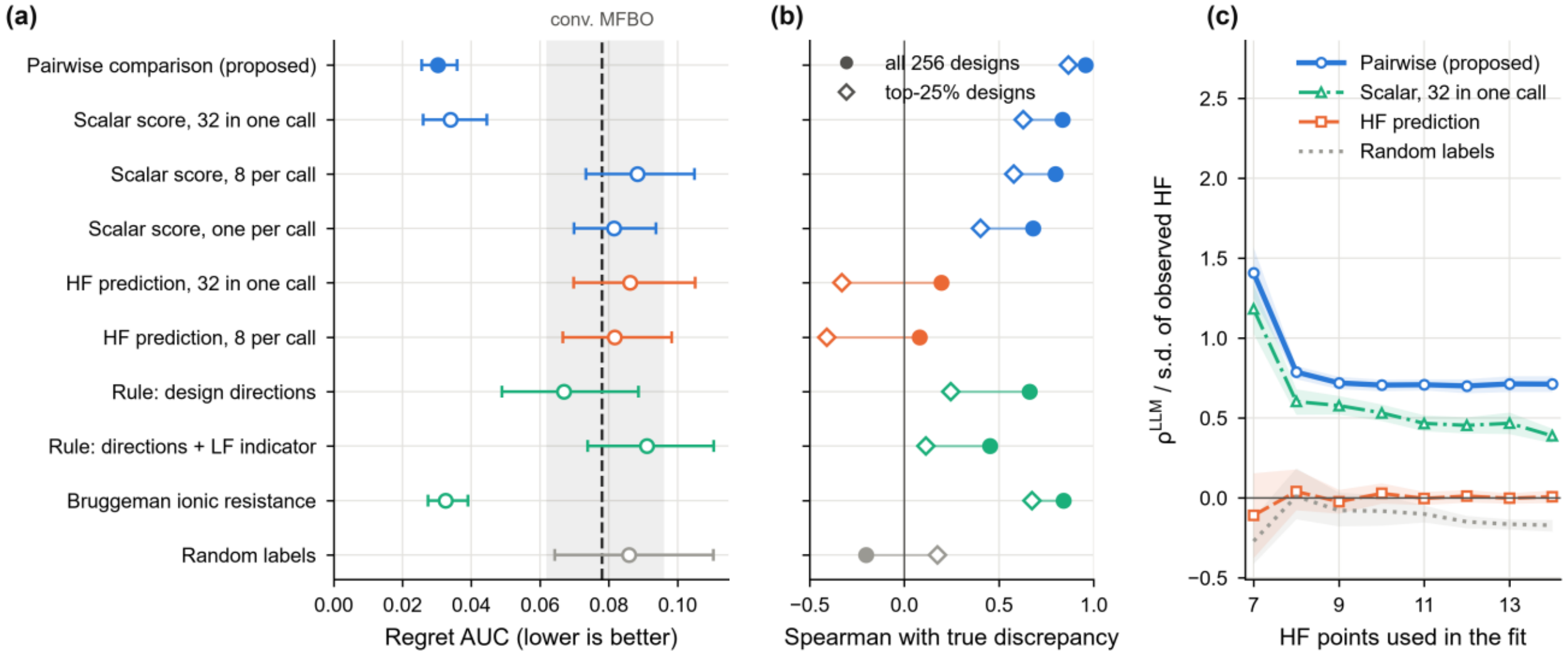

**Fig. E5.** Results for different query targets and formats (battery problem, 20 seeds): (a) mean regret AUC for each information source of the knowledge basis with 95% bootstrap CIs (dashed line and shading: conventional MFBO); (b) Spearman rank correlation between the knowledge basis and the true residual discrepancy (circles: all 256 designs; diamonds: top 25% of designs); and (c) mean $\rho^{\text{LLM}}$ calibrated with HF data and divided by the SD of the observed HF values, with 95% CIs.

**Table E1.** Results for different information sources of the knowledge basis (battery, 20 seeds). Basis accuracy is the Spearman rank correlation with the true residual discrepancy computed post hoc on 256 designs (not used during BO).

| Query target | Information source of the knowledge basis | LLM calls | Regret AUC ↓ | Gain over MFBO (seeds won) | Basis accuracy (all / top 25%) |
|---|---|---|---|---|---|
| – | Conventional MFBO (no basis) | 0 | 0.0781 | – | – |
| **Residual discrepancy** | **Pairwise comparison (proposed)** | **24** | **0.0303** | **+0.048 (18/20)** | **0.96 / 0.87** |
| Residual discrepancy | Scalar score, all 32 at once | 1 | 0.0340 | +0.044 (18/20) | 0.83 / 0.63 |
| Residual discrepancy | Scalar score, 8 at a time | 4 | 0.0883 | −0.010 (6/20) | 0.80 / 0.58 |
| Residual discrepancy | Scalar score, one at a time | 32 | 0.0814 | −0.003 (6/20) | 0.68 / 0.40 |
| HF value | Numerical prediction, all 32 at once | 1 | 0.0862 | −0.008 (7/20) | 0.19 / −0.33 |

| Query target | Information source of the knowledge basis | LLM calls | Regret AUC ↓ | Gain over MFBO (seeds won) | Basis accuracy (all / top 25%) |
|---|---|---|---|---|---|
| HF value | Numerical prediction, 8 at a time | 4 | 0.0817 | −0.004 (7/20) | 0.08 / −0.41 |
| (No LLM) | Rule: design-variable directions | 0 | 0.0669 | +0.011 (14/20) | 0.66 / 0.24 |
| (No LLM) | Rule: directions + LF transport indicator | 0 | 0.0910 | −0.013 (5/20) | 0.45 / 0.11 |
| (No LLM) | Quantitative physical indicator: Bruggeman ionic resistance | 0 | 0.0325 | +0.046 (17/20) | 0.84 / 0.67 |
| (Negative control) | Random labels | 0 | 0.0858 | −0.008 (11/20) | −0.20 / 0.18 |

**Results.** As shown in Fig. E5 and Table E1, the conditions significantly better than conventional MFBO were the pairwise comparison, the scalar score with all 32 designs at once, and the Bruggeman physical indicator (Wilcoxon signed-rank test, $p \leq 10^{-4}$), and among them the pairwise comparison had the highest basis accuracy.

- **Query target:** Numerical predictions of HF values gave no clear gain over conventional MFBO, regardless of how the calls were organized. This is interpreted as resulting from the predictions combining the level information already provided by the LF responses with the discrepancy, leaving little information to add to the LF basis.
- **Query format:** Scalar scores worked only when all 32 designs were scored at once. When scored 8 at a time, the mean score of each call stayed at 0.45–0.49 regardless of the true discrepancy, meaning that the scale was reset in every call; when scored one at a time, the scores were compressed and the basis accuracy dropped to 0.68. Pairwise comparisons do not require a common scale and thus avoid this problem.
- **Rule bases:** Rules that translated P1 directly gave no gain (+0.011 with design-variable directions only and −0.013 with the LF transport indicator added). The Bruggeman indicator, which adds a functional form and exponents absent from P1, gave a gain (+0.046), but its basis accuracy (0.84; 0.67 for the top 25%) was lower than that of the LLM basis, which used only qualitative descriptions (0.96 and 0.87).
- **Controls and coefficients:** Random labels neither helped nor hurt (11/20 seeds), and labels with all directions reversed produced the same trajectories with only the sign of the coefficient reversed (Section 3.3.3). The coefficient of the pairwise-comparison basis remained positive in all 20 seeds, whereas the coefficients of uninformative bases stayed near zero (Fig. E5(c)).
- **Sensitivity to the global scale:** For $\rho \in [0.55, 1.0]$, the rank correlation between the basis and $d_\rho$ barely changed (0.94–0.96), and all 13 A/B judgments agreed with the true order over the entire range. The pairs on which the LLM abstained had much smaller differences in true discrepancy than the A/B pairs (median 24 vs. 128 Wh/L), and 4 of the 11 abstained pairs reversed their order within this range.

## E.5 Pair-selection strategies

To examine the effect of active pair querying (Eq. (6)), the query budget in the battery problem was fixed at $B = 24$, and only the pair-selection criterion was changed. The alternative criteria are (i) BALD, which maximizes the mutual information of the label, and (ii) maximum difference variance, which selects the pair with the largest $v_{ab}$. The protocol is the same as in Section 4.4 (Sobol initial design with 7 points, GPT-6.1 Sol with high reasoning effort, P1), but the 24 queries of each strategy were generated anew; the result of active pair querying therefore comes from a query plan independent of that for P1 in Table 3.

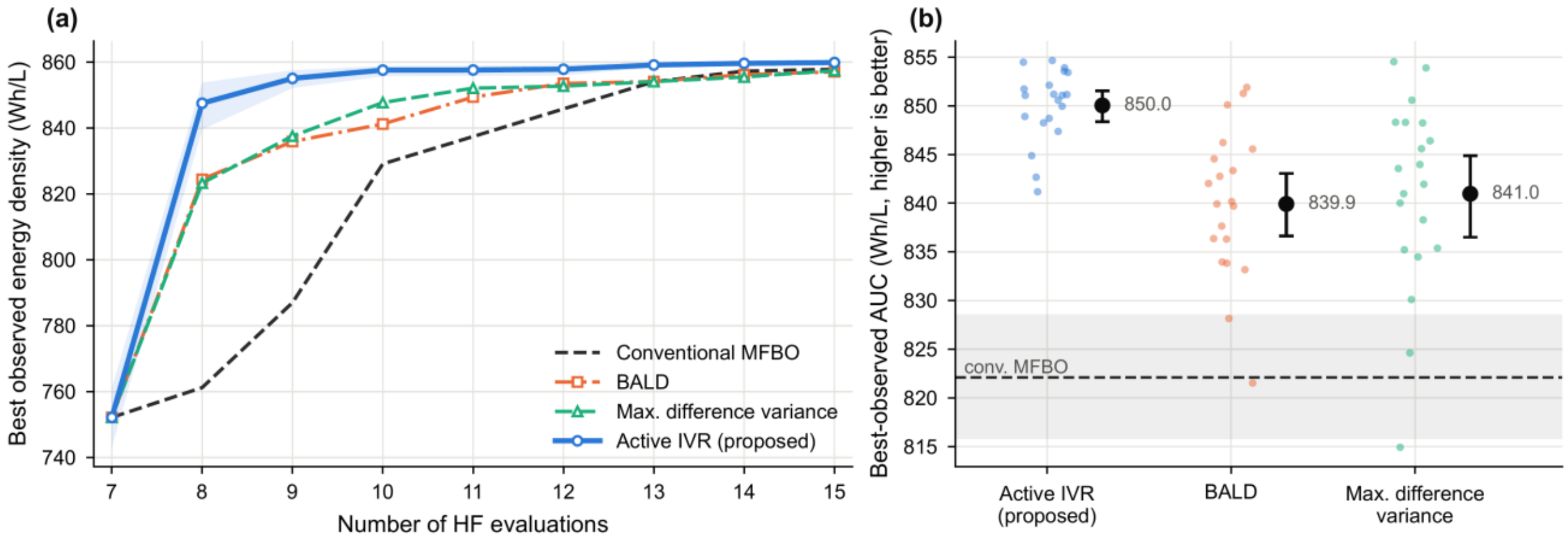

**Fig. E6.** Comparison of pair-selection strategies (battery problem, query budget 24, 20 seeds): (a) mean best-observed DFN objective for each strategy (shading: 95% bootstrap CI of the proposed method); and (b) seed-wise best-observed AUC (dots) with 95% CIs of the mean (dashed line and shading: conventional MFBO).

**Table E2.** Results of the pair-selection strategies (battery, query budget 24, 20 seeds). The best-observed AUC is in Wh/L; higher is better.

| Strategy | Mean best-observed AUC (SD) | Valid comparisons / abstentions |
|---|---|---|
| **Active pair querying (proposed)** | **850.05 (3.74)** | **12 / 12** |
| BALD | 839.93 (7.68) | 9 / 15 |
| Maximum difference variance | 840.97 (9.87) | 8 / 16 |
| Conventional MFBO | 822.11 (15.32) | – |
| HF-only BO | 808.88 (34.51) | – |

**Results.** As shown in Fig. E6 and Table E2, all three criteria achieved a higher mean performance than conventional MFBO, and active pair querying achieved the highest mean and the smallest seed-to-seed variation. The AUC gain of active pair querying over conventional MFBO was 27.9 Wh/L (95% CI [21.8, 34.1], 19/20 seeds). BALD and maximum difference variance focus on the uncertainty of individual labels and selected many ambiguous pairs on which the LLM abstained (15–16 abstentions), so fewer queries yielded valid comparisons. In contrast, active pair querying is based on the variance reduction of the entire knowledge basis and therefore achieved a higher fraction of valid comparisons (12/24).

**Reproducibility of independent query plans.** Because LLM responses are stochastic, query results can differ even under the same conditions. The two independent query plans with the same prompt (P1), LLM (GPT-6.1 Sol, high reasoning effort), and protocol, namely that of P1 in Table 3 and that of active pair querying in Table E2, yielded best-observed AUCs of 846.9 Wh/L (15 valid comparisons) and 850.0 Wh/L (12 valid comparisons), a difference of 3.1 Wh/L. Both exceeded conventional MFBO (822.1 Wh/L) by 24.8–27.9 Wh/L under a protocol whose initial design and reasoning effort differ from those in Section 4.2.